\documentclass[10pt,twocolumn,twoside]{IEEEtran}

\ifCLASSINFOpdf
\else
\fi

\usepackage{cite}
\usepackage{amsmath,mathtools,amsfonts,amsthm} 
\usepackage{amssymb}
\usepackage{bbm}
\usepackage{verbatim}
\usepackage{graphicx}  
\usepackage{enumerate}
\usepackage[font=small]{caption}
\usepackage{color}
\usepackage{subcaption}
\usepackage{tikz}
\usetikzlibrary{calc,positioning,shapes,shadows,arrows,fit}
\usepackage[color=yellow, textsize=small, textwidth=\marginparwidth, draft]{todonotes}

\usepackage{mathtools}
\usepackage{etoolbox}
\preto\subequations{\ifhmode\unskip\fi}
\usepackage{mathrsfs}
\newcommand{\colvec}[2][.9]{%
  \scalebox{#1}{%
    \renewcommand{\arraystretch}{1}%
    $\begin{bmatrix}#2\end{bmatrix}$%
  }
}

\usepackage{blkarray, bigstrut}

\newtheorem{corollary}{Corollary}
\newtheorem{lemma}{Lemma}
\newtheorem{proposition}{Proposition}

\newtheorem{definition}{Definition}
\newtheorem{theorem}{Theorem}

\newcommand{\scr}{\scriptscriptstyle}

\begin{document}
	
	\title{Cooperative Manipulation via Internal Force Regulation: A Rigidity Theory Perspective}
		
	\author{Christos K. Verginis,
		Daniel Zelazo,
		and {Dimos V. Dimarogonas}
		\thanks{C. K. Verginis is with the Division of Signals and Systems, Department of Electrical Engineering, Uppsala University, Uppsala, Sweden. Email: {\tt\small christos.verginis@angstrom.uu.se}. D. V. Dimarogonas is with 
		KTH Royal Institute of Technology, Stockholm, Sweden. Email: {\tt\small dimos@kth.se}. D. Zelazo is with 
		Technion-Israel  Institute  of  Technology,  Haifa. Email: {\tt\small  dzelazo@technion.ac.il}. 
	}
		}

	%

		\maketitle

	\begin{abstract}
		This paper considers the integration of rigid cooperative manipulation with rigidity theory. Motivated by rigid models of cooperative manipulation systems, i.e., where the grasping contacts are rigid, we introduce first the notion of bearing and distance rigidity for graph frameworks in $\mathsf{SE}(3)$. Next, we associate the nodes of these frameworks to the robotic agents of rigid cooperative manipulation schemes and we express the object-agent internal  forces by using the graph rigidity matrix, which encodes the infinitesimal rigid body motions of the system. Moreover, we show that the associated cooperative manipulation grasp matrix is related to the rigidity matrix via a range-nullspace relation, based on which we provide novel results on the relation between the arising interaction and internal forces and consequently on the energy-optimal force distribution on a cooperative manipulation system. Finally, simulation results enhance the validity of the theoretical findings.

	\end{abstract}
	
	\begin{IEEEkeywords}
		Cooperative manipulation, infinitesimal rigidity, distance rigidity, bearing rigidity.
	\end{IEEEkeywords}

	\IEEEpeerreviewmaketitle

	\section{Introduction}
	\IEEEPARstart{M}{ulti}-robot systems have received a considerable amount of attention during the last decades, due to the advantages they offer with respect to single-robot setups. 
	Example problems 
	include 
	consensus/rendezvous, connectivity maintenance, formation control, 
	and robotic manipulation. In the latter case, 
	multi-robot frameworks can yield significant advantages due to the potentially heavy payloads or challenging maneuvers.
	This work focuses on bridging the fields of cooperative robotic manipulation and robot formation control by associating the inter-agent interaction forces of the first to inter-agent geometric relations of the latter.
	
	The goal of robot formation control is to control each robot using local information from neighboring agents so that the entire team forms a desired spatial geometric pattern \cite{oh2015survey}. A special instance of formation control with  
	numerous applications 
	is 
	\textit{rigid formations}. Two cases of rigid formation control have been widely studied in the literature, namely \textit{distance rigidity} and \textit{bearing rigidity}.  Rigidity theory, a branch of discrete mathematics, explores under what conditions can the geometric pattern of a network be determined given that the length (distance) or bearing of each edge in a network of nodes is fixed.  This theory has been applied in 
	distance and bearing formation control and localization problems \cite{Automatica_formation_18,de2016distributed,chen2017global,zelazo2015decentralized,tian2013global,zhao2019bearingRig,Tron15RigidComp}.
	
	In this paper, we introduce the notion of \textit{distance and bearing rigidity}, which studies under what conditions can the geometric pattern of a multi-agent system be uniquely determined if both the \textit{distance} and the \textit{bearing} of each edge is fixed. Moreover, we combine the latter with \textit{rigid} cooperative manipulation, i.e., configurations where a number of robots carry an object via rigid contact points. 
	
	{Cooperative manipulation is a special form of constrained dynamical systems \cite{lin2018projected,aghili2005unified,erhart2015internal,Udwadia92NewPerspective,Udwadia93Constrained,donner2018physically}}. The majority of related works assume that the robotic agents are attached to the object via \textit{rigid grasps}, and hence the overall system can be 
	considered as a closed-chain robotic agent. In terms of control design, most  works consider decentralized schemes, where there is no communication between the agents, and use impedance and/or force control \cite{khatib1996decentralized,ficuciello2014cartesian,yoshikawa1993coordinated,erhart2013impedance}, possibly with contact force/torque measurements (e.g., \cite{tsiamis2015cooperative,heck2013internal}). In addition, numerous works consider unknown dynamics/kinematics of the agents and the object and/or external disturbances \cite{verginis2019Robust,erhart2013adaptive,ponce2016cooperative,marino2017distributed}.
	
	An important property in rigid cooperative manipulation systems 
	that has been studied thoroughly in the {related} literature is the regulation of internal forces. Internal forces are forces exerted by the agents at the grasping points that do not contribute to the motion of the object. While a certain amount of such forces is required in many cases (e.g., to avoid contact loss in multi-fingered manipulation), they need to be minimized in order to prevent object damage and unnecessary effort of the agents. Most works in rigid cooperative manipulation assume a certain decomposition of the {interaction} forces in motion-inducing and internal ones, without explicitly showing that the actual internal forces will be indeed regulated to the desired ones (e.g., \cite{tsiamis2015cooperative,heck2013internal}); \cite{walker1991analysis,williams1993virtual,chung2005analysis,erhart2015internal,donner2018physically} analyze specific load decompositions based on whether they provide internal force-free expressions, whereas \cite{erhart2016model} is concerned with the cooperative manipulation interaction dynamics.	
	The decompositions in the aforementioned works, however, are based on the inter-agent distances and do not take into account the actual dynamics of the agents. The latter, as we show in this paper, are tightly connected to the internal forces as well as their relation to the total force exerted by the agents. 
	
	{More specifically, the contribution of this paper is twofold. 
	Firstly, we 
	integrate rigid cooperative manipulation with rigidity theory. Motivated by rigid cooperative manipulation systems, where the inter-agent distances \textit{and} bearings are fixed, we introduce the notion of \textit{distance and bearing rigidity} in the special Euclidean group $\mathsf{SE}(3)$. Based on recent results, we show next that the internal forces in a rigid cooperative manipulation system, {consisting of more than 2 robotic agents}, depend on the distance and bearing rigidity matrix, a matrix that encodes the allowed coordinated motions of the multi-agent-object system. Moreover, we prove that the cooperative manipulation grasp matrix, which relates the object and agent velocities, is connected via a range-nullspace relation to the rigidity matrix.
	Secondly, we rely on the aforementioned findings to provide new results on the internal force-based rigid cooperative manipulation. We {derive}
	novel results on the relation between the arising interaction and internal forces in a cooperative manipulation system.  This leads to novel conditions on the internal force-free object-agents force distribution and consequently to optimal, in terms of energy resources, cooperative manipulation. {Bearing rigidity has been used before in \cite{briot2019physical} to analyze the properties of virtual closed-loop mechanisms and parallel robots; similarly,
	our analysis provides new physical insights for the fields of cooperative manipulation and rigidity theory.} 
	This paper extends our preliminary conference version \cite{verginis_submitted}, which tackles optimal cooperative manipulation by regulating the internal forces. That work, however, does not associate cooperative manipulation with rigidity theory or provide \textit{explicit} results on the optimal object-agents force distribution.  }
	
	The rest of the paper is organized as follows. Section \ref{sec:preliminaries} provides notation and necessary background. Section \ref{sec:Coop manip model} provides the cooperative manipulation model and Section \ref{sec:Rigidity} discusses distance and bearing rigidity. The main results of the paper are given in Section \ref{sec:Main results}, and Section \ref{sec:discussion} discusses features of our analysis. Finally, Section \ref{sec:Sim/Exp results} presents simulation results and Section \ref{sec:Conclusion+FW} concludes the paper. 
	
%
		
	\section{Preliminaries} \label{sec:preliminaries}

	
	The set of positive integers is denoted by $\mathbb{N}$ and the real $n$-coordinate space, with $n\in\mathbb{N}$, by $\mathbb{R}^n$.
    The $n\times n$ identity matrix is denoted by $I_n$, the $n$-dimensional zero vector by $0_n$ and the $n\times m$ matrix with zero entries by $0_{n\times m}$. We write $0$ instead of $0_n$ when $n$ is clear from the context. The vectors of the canonical basis of $\mathbb{R}^d$ are indicated as 
$\mathbf{e}_i, \; i \in \{1 \ldots d\}$, and they have a one in the $(i\, \text{mod}\, d)$-th entry  and zeros elsewhere.
	Given a matrix $A\in\mathbb{R}^{n\times m}$, we use $A^\dagger$ for its Moore-Penrose inverse, {and null($A$), range($A$) for its nullspace and range space, respectively}.  
	For a discrete set $\mathcal{N}$, $|\mathcal{N}|$ denotes its cardinality. 
	Given $a,{b}\in\mathbb{R}^3$, $S(a)$ is the skew-symmetric matrix
	defined according to $S(a)b = a\times b$. 
	 In addition, $ \mathsf{S}^n$ denotes the $(n+1)$-dimensional sphere and $\mathsf{SO}(3)$ $\mathsf{SE}(3)$ the rotation and special Euclidean group, respectively; $P_r(x) \coloneqq I_n - \frac{xx^\top}{\|x\|^2}$ projects vector
	 $x\in\mathbb{R}^n$ onto the orthogonal complement of $x$, {i.e., the subspace $\{y\in\mathbb{R}^n:y^\top x = 0\}$}.
			A graph $\mathcal{G}$ is a pair $(\mathcal{N},\mathcal{E})$, where $\mathcal{N}$ is a finite set of $N= |\mathcal{N}|\in\mathbb{N}$ nodes, and $\mathcal{E} \subseteq \mathcal{N}\times \mathcal{N}$ is a finite set of $|\mathcal{E}|$ edges.
	The complete graph on $N$ nodes is denoted by $\mathcal{K}_N$. {All vectors and vector differentiations are expressed with respect to a known inertial reference frame, unless otherwise stated.}

We also make use of some properties from linear algebra.
{1)} For any matrix $H$, it holds that $H^\dagger = H^\top (HH^\top)^\dagger$ \cite[Theorem 3.8]{albert72Pseudoinverse}. {2)} For a matrix $A \in \mathbb{R}^{n\times m}$, and $B \coloneqq K A$, where $K\in\mathbb{R}^{n\times n}$ is an invertible matrix,  it holds that $A^\dagger A = B^\dagger B$. {3)} Let $A, B \in \mathbb{R}^{n\times m}$ such that $\textup{range}(A^\top) = \textup{null}(B)$. Then it holds that $A^\dagger A + B^\dagger B = I_m$. {4)}  For matrices $A,B\in \mathbb{R}^{n\times m}$, $A$ is \emph{left equivalent} (or \emph{row equivalent}) to $B$ if and only if there exists an invertible matrix $P\in\mathbb{R}^{n\times n}$ such that $A= PB$. It then can be shown that $A$ and $B$ are  left-equivalent if and only if $\textup{null}(A) = \textup{null}(B)$.

    { 
    \begin{lemma}[Gauss' principle \cite{Udwadia92NewPerspective,Udwadia93Constrained}] \label{lem:gauss}
    Let an unconstrained system described by the configuration variables $q\in\mathbb{R}^n$ and evolving according to $M(q,t)\ddot{q} = Q(q,\dot{q},t)$ where $M\in\mathbb{R}^{n\times n}$ is positive definite. 
    Assume now that the system is subjected to $m$ consistent constraints of the form $A(q,\dot{q},t)\ddot{q} = b(q,\dot{q},t)$. Then, {the acceleration $\ddot{q}$ of the constrained system} is given by the constrained minimization problem 
    \begin{align*}
					&\min_{\ddot{q}}\ \ [\ddot{q} - \alpha]^\top M(q) [\ddot{q} - \alpha] 	
					&\hspace{1mm} \textup{s.t.} \ \ \ A(q,\dot{q},t)\ddot{q} = b(q,\dot{q},t),
			\end{align*}
			where $\alpha \coloneqq M(q)^{-1} Q(q,\dot{q},t)$ is the acceleration of the unconstrained system.
    \end{lemma}}
   
    {

	\section{Cooperative Manipulation Modeling} \label{sec:Coop manip model}
	
	We provide in this section the dynamic modeling of the rigid cooperative manipulation system. {A key feature of the model is the grasp matrix, which, as will be clarified,
	motivates the introduction of the notion of distance and bearing rigidity in the next section and the association between the two.}
		
	{Consider $N$ robotic agents, indexed by the set $\mathcal{N} \coloneqq \{1,\dots,N\}$, rigidly grasping an object.}
	We denote by $q_i,\dot{q}_i \in\mathbb{R}^{n_i}$, with $n_i\in\mathbb{N}, \forall i\in\mathcal{N}$, the generalized joint-space variables and their  derivatives of agent $i$.
	The overall joint configuration is then $q \coloneqq [q^\top _1,\dots,q^\top _N]^\top , \dot{q} \coloneqq [\dot{q}^\top _1,\dots,\dot{q}^\top _N]^\top \in\mathbb{R}^{n}$, with $n \coloneqq \sum_{i\in\mathcal{N}}n_i$. In addition, we denote the position and rotation matrix of the $i$th end-effector by  $p_i\in\mathbb{R}^3$ {and $R_i \in \mathsf{SO}(3)$, respectively}. 
	Similarly, the velocity of the $i$th end-effector is denoted by $v_i \coloneqq [\dot{p}_i^\top, \omega^\top_i]^\top$, where $\omega_i\in\mathbb{R}^3$ is the respective angular velocity, and it holds that $v_i = J_i(q_i) \dot{q}_i$, where $J_i:\mathbb{S}_i \to \mathbb{R}^{6\times n_i}$ is the robot Jacobian, and $\mathbb{S}_i \coloneqq \{q_i\in\mathbb{R}^{n_i} : \dim(\textup{null}(J_i(q_i))) = 0\}$ is the set away from kinematic singularities \cite{siciliano2010robotics}, $\forall i\in\mathcal{N}$. Moreover 
	we denote $x_i \coloneqq (p_i,R_i)\in\mathsf{SE}(3)$ and $x \coloneqq (x_1,\dots,x_N) \in \mathsf{SE}(3)^N$.
	The 
	{task-space} dynamics of the agents are \cite{siciliano2010robotics}:
	\begin{align}
		{M}({x})\dot{v}+{C}({x},\dot{{x}})v + {g}({x}) &= {u}-h,\label{eq:manipulator dynamics_task space vector_form}
	\end{align}
	{where 
		$v \coloneqq [v^\top _1,\dots,v^\top _N]\in\mathbb{R}^{6N}$, $h \coloneqq [h^\top _1,\dots,h^\top _N]^\top$, $u \coloneqq [u^\top _1,\dots,u^\top _N]^\top $, $g \coloneqq [g^\top _1,\dots,g^\top _N]^\top \in \mathbb{R}^{6N}$; the terms 
		$M_i :\mathsf{SE}_i\to\mathbb{R}^{6\times 6}$, $C_i : \mathsf{SE}_i\times\mathbb{R}^{6}\to\mathbb{R}^{6\times 6}$, $g_i :\mathsf{SE}_i\to\mathbb{R}^6$ are their positive definite inertia, Coriolis, and gravity terms, respectively, which are well-defined when $q_i \in \mathbb{S}_i$, $i\in\mathcal{N}$; $h_i\in\mathbb{R}^{6}$ {are the forces between the object and the agents}, and  $u_i\in\mathbb{R}^6$
		the task-space inputs, $\forall i\in\mathcal{N}$.}

	Regarding the object, we denote by {$x_{\scriptscriptstyle O}\coloneqq (p_{\scriptscriptstyle O},R_{\scr O}) \in \mathsf{SE}(3)$}, 
	$v_{\scriptscriptstyle O} \coloneqq [\dot{p}^\top_{\scriptscriptstyle O}, \omega^\top _{\scriptscriptstyle O}]^\top \in\mathbb{R}^{12}$ the pose and generalized velocity of the object's center of mass.
	We consider the following second order dynamics, which can be derived based on the Newton-Euler formulation: 
	\begin{subequations} \label{eq:object dynamics}
		\begin{align}
		& \dot{R}_{\scriptscriptstyle O} = S(\omega_{\scr O}) R_{\scr O}  \label{eq:object dynamics 1}\\
		& M_{\scriptscriptstyle O}(x_{\scriptscriptstyle O})\dot{v}_{\scriptscriptstyle O}+C_{\scriptscriptstyle O}(x_{\scriptscriptstyle O},\dot{x}_{\scriptscriptstyle O})v_{\scriptscriptstyle O}+g_{\scriptscriptstyle O}(x_{\scriptscriptstyle O}) = h_{\scriptscriptstyle O}, \label{eq:object dynamics 2}
		\end{align}
	\end{subequations}
	where $M_{\scriptscriptstyle O}:\mathbb{M}\to\mathbb{R}^{6\times6}$ is the positive definite inertia matrix, $C_{\scriptscriptstyle O}:\mathbb{M}\times\mathbb{R}^{6}\to\mathbb{R}^{6\times6}$ is the Coriolis matrix, $g_{\scriptscriptstyle O}:\mathbb{M}\to\mathbb{R}^6$ is the gravity vector, and $h_{\scriptscriptstyle O}\in\mathbb{R}^6$ is the vector of generalized forces acting on the object's center of mass. 
	
	
	\begin{figure}
		\centering
		\includegraphics[width = 0.25\textwidth]{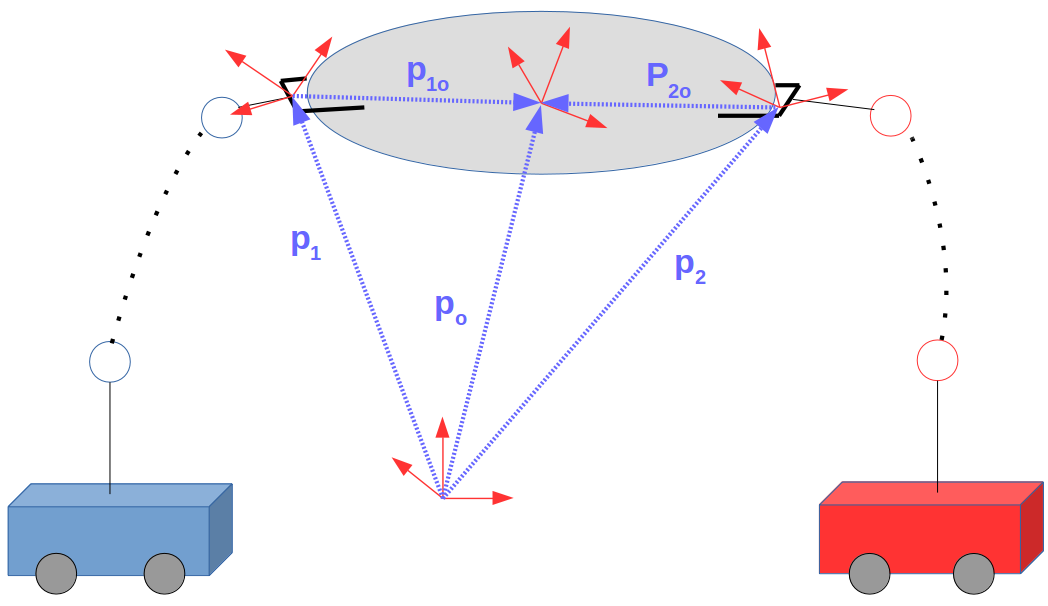}
		\caption{Two robotic agents rigidly grasping an object.\label{fig:Two-robotic-arms}}
		\vspace{-.35cm}
	\end{figure}
	
	In view of Fig. \ref{fig:Two-robotic-arms} and the grasping rigidity, one obtains \cite{verginis2019Robust}
	\begin{equation}
	v_i=J_{\scriptscriptstyle O_i}(x_i) v_{\scriptscriptstyle O}, \ \forall i\in\mathcal{N}, \label{eq:J_o_i}
	\end{equation}
	where $J_{\scriptscriptstyle O_i}:\mathsf{SE}(3)\to\mathbb{R}^{6\times6}$ is the object-to-agent Jacobian matrix, with
	\begin{equation}
	J_{\scriptscriptstyle O_i}(x_i) = \left[\begin{array}{cc}
	I_{3} & -S(p_{i\scr O})\\
	0_{3\times 3} & I_{3}
	\end{array}\right], \label{eq:J_o_i_def}
	\end{equation}
	and $p_{i \scr O} \coloneqq p_i - p_{\scr O}$, $\forall i\in\mathcal{N}$;
	$J_{\scr O_i}$ is always full-rank, due to the rigidity of the grasping contacts. {The grasp matrix is formed by stacking $J_{\scr O_i}^\top$ as 
	\begin{equation} \label{eq:grasp matrix def.}
		G(x) \coloneqq [J_{\scriptscriptstyle O_1}(x_1)^\top,\dots,J_{\scriptscriptstyle 	O_N}(x_N)^\top] \in \mathbb{R}^{6 \times 6N},
	\end{equation}
	 and has full column rank due to the rigidity of the grasping contacts; \eqref{eq:J_o_i} can now be written in stack vector form as 
	 \begin{equation} \label{eq:grasp matrix velocities}
	 	v = G(x)^\top v_{\scr O}.
	 \end{equation}}	
	The kineto-statics duality {\cite{siciliano2010robotics}} along with the grasp rigidity suggest that the force $h_{\scriptscriptstyle O}$ acting on the object's center of mass and the generalized forces $h_i,i\in\mathcal{N}$, exerted by the agents at the grasping points, are related through:
	\begin{equation}
	h_{\scriptscriptstyle O}=G(x) h.\label{eq:grasp matrix}
	\end{equation}	
	By using \eqref{eq:manipulator dynamics_task space vector_form}, \eqref{eq:grasp matrix}, and \eqref{eq:object dynamics}, we obtain the coupled dynamics:
	\begin{equation}
	{	{M}_\mathsf{c}(\bar{x})\dot{v}_{\scriptscriptstyle O}+{C}_\mathsf{c}(\bar{x},\dot{\bar{x}})v_{\scriptscriptstyle O}+{g}_\mathsf{c}(\bar{x})  = G(x) u,\label{eq:coupled dynamics} }
	\end{equation}
	where {$M_\mathsf{c}  \coloneqq M_{\scriptscriptstyle O}+ G M G^\top$,
		$C_\mathsf{c} \coloneqq C_{\scriptscriptstyle O}+ G C G^\top + G M \dot{G}^\top$,
		$g_\mathsf{c}  \coloneqq g_{\scriptscriptstyle O} + G g$, 
		$\bar{x}$ is the coupled state $\bar{x}\coloneqq [x^\top,x^\top_{\scriptscriptstyle O}]^\top\in \mathsf{SE}(3)^{N+1}$}, and we have omitted the arguments for brevity.
	{The interaction forces $h$ between the object and the agents} can be decoupled into motion-induced and internal forces 
	\begin{equation} \label{eq:interaction forces}
	h = h_\text{m} + h_\text{int}.
	\end{equation}
	The internal forces $h_\text{int}$ are squeezing forces that the agents exert to the object and belong to the nullspace of $G(x)$ (i.e., $G(x) h_\textup{int} = 0$). {Intuitively, when $h=h_{\textup{int}}$, it holds that $G(x)(u - M\dot{v} - C v -g) = 0$
	and the object moves according to $h_{\scr O}=M_{\scr O}\dot{v}_{\scr O} + C_{\scr O} v_{\scr O} + g_{\scr O} =  0$.}
	Hence, $h_\text{int}$ does not contribute to {the} object's motion and results in internal stresses that might damage it. {An} analytic expression for $h$ and $h_\text{int}$ is given in Section \ref{sec:Main results}.
	
	{Note from \eqref{eq:grasp matrix velocities} that the agent velocities $v$ belong to the range space of $G(x)^\top$. Therefore, since $G(x)$ is a matrix that encodes rigidity constraints, this motivates the association of $G(x)$ to the \emph{rigidity matrix} used in formation rigidity theory, and of the rigid cooperative manipulation scheme to a multi-agent rigid formation scheme. {To this end, we introduce next in Section \ref{sec:Rigidity} the notion of Distance and Bearing Rigidity. In Section \ref{sec:forces+rigid}, we connect the latter with the cooperative manipulation system \eqref{eq:coupled dynamics}, and in Section \ref{sec:manip internal} we use this connection to derive new results on  cooperative manipulation free from internal forces.
	We argue that the association of rigid cooperative manipulation with rigidity theory, which has not been considered before, provides new physical insights in the intersection of the two fields (illustrated in Theorem \ref{th:null G range R_T} of Section \ref{sec:Main results}). 
	It also paves the way for drawing novel links that could help solve problems of one by leveraging the rich literature of the other.}
	}

	\section{Distance and Bearing Rigidity in $\mathsf{SE}(3)$} \label{sec:Rigidity}
	


	\textcolor{black}{
		We begin by recalling that the range space of the grasp matrix $G(x)^T$ corresponds to the rigid body translations and rotations of the system.  While $G$ appears naturally in the context of dynamic modeling of rigid bodies, it is also indirectly related to the notion of structural rigidity in discrete geometry.
		}
	
	\textcolor{black}{	
		In the classical structural rigidity theory, one considers a collection of rigid bars connected by joints allowing free rotations around the joint axis (\emph{bar-and-joint} frameworks).  One is then interested in understanding what are the allowable motions of the framework, i.e., those motions that preserve the lenghts of the bars and their connections to the joints.  The so-called \emph{trivial motions} for these frameworks are precisely the rigid body translations and rotations of the system.  For some frameworks, there may be additional motions, known as \emph{flexes}, that also preserve the constraints.  This is captured by the notion of \emph{infinitesimal motions} of the framework and is characterized by the \emph{rigidity matrix} of the framework \cite{asimow1978rigidity}. 
	}
	
	\textcolor{black}{Since, in a rigid cooperative manipulation system, the relative distances and bearings among the agents and the object are \textit{fixed}, we naturally consider frameworks that encode both the lengths of bars and pose of the joints, 
	leading to a \emph{distance and bearing}-type framework.  This relates notions from distance rigidity, and recent works in bearing rigidity for frameworks  embedded in $\mathsf{SE}(2)$ and $\mathsf{SE}(3)$\cite{bearingSE2,bearingSE3, Michieletto2021}.
	}	
	\textcolor{black}{In this direction, we introduce 
	the concept of distance and bearing rigidity (abbreviated as D\&B Rigidity 
	). }
	\textcolor{black}{
	For this work we focus on the notion of infinitesimal rigidity for D\&B frameworks. We first formally define a D\&B framework in $\mathsf{SE}(3)$:}
	\begin{definition} 
		A framework in $\mathsf{SE}(3)$ is a triple $(\mathcal{G},p_\mathcal{G},R_\mathcal{G})$, where $\mathcal{G}\coloneqq (\mathcal{N},\mathcal{E})$ is a graph, 
		$p_\mathcal{G}: \mathcal{N} \to \mathbb{R}^3$ is a function mapping each node to a position in $\mathbb{R}^3$, and $R_\mathcal{G} : \mathcal{N} \to \mathrm{SO}(3)$ is a function associating each node with an orientation element of $\mathrm{SO}(3)$ (both with respect to an inertial frame).
	\end{definition}
	
	In this work we employ the Special Orthogonal Group (rotation matrices) $\{ R \in \mathbb{R}^{3\times 3} : R^\top R = I_3, \textup{det}(R) = 1 \} $ to express the orientation of the agents. Moreover, we use the shorthand notation $p_i \coloneqq p_\mathcal{G}(i)$, $R_i \coloneqq R_\mathcal{G}(i)$, $p \coloneqq [p_1^\top,\dots,p_N^\top]^\top \in\mathbb{R}^{3N}$, $R \coloneqq (R_1,\dots,R_N)\in\mathsf{SO}(3)^N$, $x_i \coloneqq (p_i,R_i)\in\mathsf{SE}(3)$, and $x \coloneqq (x_1,\dots,x_N) \in \mathsf{SE}(3)^N$.	
	

	
	{The distances and bearings in a framework can be summarized through the following \textit{$\mathsf{SE}(3)$ D\&B rigidity function}, $\gamma_\mathcal{G}$, that encodes the rigidity constraints in the framework. 
		Consider a directed graph $\mathcal{G}=(\mathcal{N},\mathcal{E})$, where $\mathcal{E} \subseteq \{ (i,j) \in \mathcal{N}^2 : i\neq j \}$, {as well as 
		$ \mathcal{E}_u \coloneqq \{(i,j)\in\mathcal{E} : i<j \} \subseteq \mathcal{E}$}. Then $\gamma_\mathcal{G}$ can be formed by the distance and bearing functions $\gamma_{e,d}:\mathbb{R}^3\times\mathbb{R}^3\to\mathbb{R}_{\geq 0}$, $\gamma_{e,b}:\mathsf{SE}(3)^2\to\mathsf{S}^2$, with 
		\begin{subequations} \label{eq:gamma 1}
			\begin{align}
			&\gamma_{e,d}(p_i,p_j) \coloneqq \frac{1}{2}\|p_i - p_j\|^2, \forall e=(i,j)\in\mathcal{E}_u, \\
			&\gamma_{e,b}(x_i,x_j) \coloneqq						
			R_i^\top \frac{p_j - p_i}{\|p_i-p_j\|},	
			\forall e=(i,j)\in\mathcal{E},
			\end{align}
		\end{subequations}
		which encodes the distance $\|p_i-p_j\|$ between two agents as well as the local bearing vector $R_i^\top \frac{p_j - p_i}{\|p_i-p_j\|}$, expressed in the frame of agent $i$. 
		Now $\gamma_\mathcal{G}$ is formed by stacking the aforementioned distance and bearing functions, i.e., $\gamma_\mathcal{G}: \mathsf{SE}(3)^N \to \mathbb{R}^{|\mathcal{E}_u|}\times\mathsf{S}^{2|\mathcal{E}|}$, with
				\vspace{-.05cm} 
		\begin{equation} \label{eq:gamma G}
		\gamma_\mathcal{G}	\coloneqq \begin{bmatrix} \gamma_d(p) \\ \gamma_b(x) \end{bmatrix} \coloneqq \begin{bmatrix}
		\gamma_{1,d},
		\dots,
		\gamma_{|\mathcal{E}_u|,d},
		\gamma_{1,b}^\top,
		\dots,
		\gamma_{|\mathcal{E}|,b}^\top
		\end{bmatrix}^\top.
		\end{equation} }
	{We have introduced the edge set $\mathcal{E}_u$ for the symmetric distance functions $\gamma_{(i,j),d} = \gamma_{(j,i),d}$ in order to avoid redundancy in the rows of $\gamma_\mathcal{G}$.}
	Note that the aforementioned expressions for $\gamma_{e,d}$, $\gamma_{e,b}$ are not unique and other choices that capture the rigidity constraints can also be made.  We also mention our slight abuse of notation, where the index $k$ in $\gamma_{k,d}$ and $\gamma_{k,b}$ refers to a labeled edge in $\mathcal{E}_u$ and $\mathcal{E}$.  
	
	{In this work, we are interested in the set of D\&B \textit{infinitesimal} motions of a framework in $\mathsf{SE}(3)$. \textcolor{black}{These can be thought as perturbations to a framework in $\mathsf{SE}(3)$ that leave $\gamma_\mathcal{G}$ unchanged.}	This set is characterized by the nullspace of the matrix appearing in the rate-of-change of $\gamma_{\mathcal{G}}$ under the kinematic equations associated with rotational motion in $\mathsf{SE}(3)$ \cite{Michieletto2021}.
	%
	That is, the nullspace of the matrix $\nabla_{(p,R)}\gamma_\mathcal{G}$, termed the $\mathsf{SE}(3)$-D\&B \emph{rigidity matrix} 
	$\mathcal{R}_\mathcal{G}:\mathsf{SE}(3)^N \to \mathbb{R}^{(|\mathcal{E}_u|+3|\mathcal{E}|) \times 6N} \coloneqq \nabla_{(p,R)}\gamma_\mathcal{G}$, i.e., 	
		{\footnotesize \begin{equation} \label{eq:rigidity matrix}
		\mathcal{R}_\mathcal{G}(x)  = \begin{bmatrix}
		\frac{\partial \gamma_{1,d}}{\partial p_1} & \frac{\partial \gamma_{1,d}}{\partial R_1} & \dots & \frac{\partial \gamma_{1,d}}{\partial p_N} & \frac{\partial \gamma_{1,d}}{\partial R_N} \\
		\vdots & & \ddots & & \vdots \\ 
		\frac{\partial \gamma_{|\mathcal{E}_u|,d}}{\partial p_1} & \frac{\partial \gamma_{|\mathcal{E}_u|,d} }{\partial R_1} & \dots & \frac{\partial \gamma_{|\mathcal{E}_u|,d} }{\partial p_N} & \frac{\partial \gamma_{|\mathcal{E}_u|,d} }{\partial R_N} \\
		\frac{\partial \gamma_{1,b}}{\partial p_1} & \frac{\partial \gamma_{1,b}}{\partial R_1} & \dots & \frac{\partial \gamma_{1,b}}{\partial p_N} & \frac{\partial \gamma_{1,b}}{\partial R_N} \\
		\vdots & & \ddots & & \vdots \\ 
		\frac{\partial \gamma_{|\mathcal{E}|,b}}{\partial p_1} & \frac{\partial \gamma_{|\mathcal{E}|,b} }{\partial R_1} & \dots & \frac{\partial \gamma_{|\mathcal{E}|,b} }{\partial p_N} & \frac{\partial \gamma_{|\mathcal{E}|,b} }{\partial R_N}
		\end{bmatrix},
		\end{equation}	}
		with
		\begin{align*}
		\frac{\partial \gamma_{e,d}}{\partial x_i} =& \begin{bmatrix}
		\frac{\partial \gamma_{e,d}}{\partial p_i}	& \frac{\partial \gamma_{e,d}}{\partial R_i}
		\end{bmatrix} = \displaystyle
		\begin{bmatrix}
		(p_i - p_j)^\top & 0_{1 \times 3} 
		\end{bmatrix}, \\    
		\frac{\partial \gamma_{e,d}}{\partial x_j} =& \begin{bmatrix}
		\frac{\partial \gamma_{e,d}}{\partial p_j}	& \frac{\partial \gamma_{e,d}}{\partial R_j}
		\end{bmatrix} = \displaystyle
		\begin{bmatrix}
		(p_j - p_i)^\top & 0_{1 \times 3} 
		\end{bmatrix}, \\   
		\frac{\partial \gamma_{e,b}}{\partial x_i} =& \begin{bmatrix}
		\frac{\partial \gamma_{e,b}}{\partial p_i}	& \frac{\partial \gamma_{e,b}}{\partial R_i}
		\end{bmatrix} = \displaystyle
		\begin{bmatrix}
		-\frac{P_r(\gamma_{e,b})}{\|p_j-p_i\|}R_i^\top &  S(\gamma_{e,b}) R_i^\top
		\end{bmatrix}, \\ 
		\frac{\partial \gamma_{e,b}}{\partial x_j} =& \begin{bmatrix}
		\frac{\partial \gamma_{e,b}}{\partial p_j}	& \frac{\partial \gamma_{e,b}}{\partial R_j}
		\end{bmatrix} =
		\begin{bmatrix}
		\frac{P_r(\gamma_{e,b})}{\|p_j-p_i\|}R_i^\top & 0_{3\times 3}
		\end{bmatrix}. 
		\end{align*}
		{The projection operator $P_r(\cdot)$ \cite{zhao2019bearingRig} is defined as in Section \ref{sec:preliminaries}.
		Infinitesimal motions, therefore, are motions $x(t)$ produced by velocities $v(t)$ that lie in the nullspace of $\mathcal{R}_\mathcal{G}$, for which it holds that $\dot{\gamma}_\mathcal{G} = \mathcal{R}_\mathcal{G}(x(t)) v(t) = 0$, where $v\coloneqq [\dot{p}_1^\top,\omega_1^\top,\dots,\dot{p}_N^\top,\omega_N^\top]^\top$, as defined in Section \ref{sec:Coop manip model}.  \textcolor{black}{The infinitesimal motions therefore depend on the number of motion degrees of freedom the entire framework possesses.  This directly relates to the structure of the underlying graph.  }
		Motions that preserve the distances and bearings of the framework for \emph{any} underlying graph are called $D\& B$ \textit{trivial motions}. This leads to the definition of \emph{infinitesimal rigidity}. 
		\begin{definition} \label{def:inf rig}
			A framework $(\mathcal{G}, p_\mathcal{G}, R_\mathcal{G})$ is D\&B infinitesimally rigid in $\mathsf{SE}(3)$ if every D\&B infinitesimal motion is a D\&B trivial motion. 
		\end{definition}
	} 

	\textcolor{black}{
		We now aim to identify 
		what the trivial motions of a D\&B framework are, and to determine conditions for a framework to be infinitesimally rigid based on properties of $\mathcal{R}_\mathcal{G}$. 
		Before we proceed, we note that the D\&B rigidity function in $\mathsf{SE}(3)$ can be seen as a superposition of the rigidity functions associated with the {classic} distance rigidity theory \cite{asimow1978rigidity} and the $\mathsf{SE}(3)$ \textcolor{black}{bearing} rigidity theory \cite{bearingSE3}.  In particular, we note that $\mathcal{R}_{\mathcal{G},d}: \mathbb{R}^{3N} \to \mathbb{R}^{|\mathcal{E}_u|\times 3N}\coloneqq \nabla_p \gamma_{d}$ is the well-studied (distance) rigidity matrix, while $\mathcal{R}_{\mathcal{G},b}:\mathsf{SE}^{3N}\to\mathbb{R}^{3|\mathcal{E}|\times6N} \coloneqq \nabla_{(p,R)}\gamma_{\mathcal{G},b}$ is the $\mathsf{SE}(3)$ \textcolor{black}{bearing} rigidity matrix.  Note that $\mathcal{R}_{\mathcal{G},d}$ 
		is associated with the framework $(\mathcal{G},p_\mathcal{G})$, which is the projection of $(\mathcal{G},p_\mathcal{G},R_\mathcal{G})$ {to} $\mathbb{R}^3$. With an appropriate permutation, $P_R$, of the columns of $\mathcal{R}_{\mathcal{G}}$, we have that 		
		\begin{align} \label{eq:R_G tilde}
		\widetilde{\mathcal{R}}_\mathcal{G} &\coloneqq   \mathcal{R}_\mathcal{G} P_R \notag \\
		&\hspace{-.6cm}= \begin{bmatrix}
		\frac{\partial \gamma_{1,d}}{\partial p_1} & \dots & \frac{\partial \gamma_{1,d}}{\partial p_N} & \frac{\partial \gamma_{1,d}}{\partial R_1} & \dots & \frac{\partial \gamma_{1,d}}{\partial R_N} \\	
		\vdots & & \ddots & & \vdots \\ 
		\frac{\partial \gamma_{M_\mathcal{G},d}}{\partial p_1} & \dots & \frac{\partial \gamma_{M_\mathcal{G},d}}{\partial p_N} & \frac{\partial \gamma_{M_\mathcal{G},d}}{\partial R_1} & \dots & \frac{\partial \gamma_{M_\mathcal{G},d}}{\partial R_N} \\    
		\frac{\partial \gamma_{1,b}}{\partial p_1} & \dots & \frac{\partial \gamma_{1,b}}{\partial p_N} & \frac{\partial \gamma_{1,b}}{\partial R_1} & \dots & \frac{\partial \gamma_{1,b}}{\partial R_N} \\
		\vdots & & \ddots & & \vdots \\ 
		\frac{\partial \gamma_{M_\mathcal{G},b}}{\partial p_1} & \dots & \frac{\partial \gamma_{M_\mathcal{G},b}}{\partial p_N} & \frac{\partial \gamma_{M_\mathcal{G},b}}{\partial R_1} & \dots & \frac{\partial \gamma_{M_\mathcal{G},b}}{\partial R_N} 
		\end{bmatrix},
		\end{align}		
		which is equal to 
		\begin{equation*}
		\widetilde{\mathcal{R}}_\mathcal{G} = \begin{bmatrix} \begin{bmatrix}\mathcal{R}_{\mathcal{G},d} &  0_{|\mathcal{E}_u|\times 3N} \end{bmatrix} \\	
		\mathcal{R}_{\mathcal{G},b} 
		\end{bmatrix} =: \begin{bmatrix} \bar{\mathcal{R}}_{\mathcal{G},d} \\	
		\mathcal{R}_{\mathcal{G},b} 
		\end{bmatrix}.
		\end{equation*}
		The nullspace of $\widetilde{\mathcal{R}}_\mathcal{G}$, therefore, is the intersection of the nullspaces of $\bar{\mathcal{R}}_{\mathcal{G},d}$ and $\mathcal{R}_{\mathcal{G},b}$.
	}
	
	\textcolor{black}{With the above interpretation, we can now understand the trivial motions to be the intersection of trivial motions associated to distance rigidity with those associated to $\mathsf{SE}(3)$ bearing rigidity.  In particular, let 
		$$\mathcal{S}_d \coloneqq \mathrm{span} \left\{ 1_N \otimes I_3, \mathcal{L}^{\circlearrowright}_{\mathbb{R}^3}(\mathcal{G})\right\},$$
		denote the trivial motions associated to a distance framework {\cite{asimow1978rigidity}}.  That is, $1_N \otimes I_3$ represents translations of the entire framework, and $\mathcal{L}^{\circlearrowright}_{\mathbb{R}^3}(\mathcal{G})$ is the rotational subspace induced by the graph $\mathcal{G}$ in $\mathbb{R}^3$, i.e., 
		$$\mathcal{L}^{\circlearrowright}_{\mathbb{R}^3}(\mathcal{G}) = \mathrm{span}\left\{ 
		\left(  I_N \otimes S(\mathbf{e}_{h})\right)p_\mathcal{G}, h=1,2,3\right\}.$$
		These motions can be produced by {the} \emph{linear} velocities of the agents.  It is known that $\mathcal{S}_d \subseteq \mathrm{null}(\mathcal{R}_{\mathcal{G},d})$ for any underlying graph $\mathcal{G}$ \cite{asimow1978rigidity}. For the matrix $\bar{\mathcal{R}}_{\mathcal{G},d}$, we can define the corresponding set
		$$\bar{S}_d \coloneqq \mathrm{span}\left\{\begin{bmatrix}
		1_N \otimes I_3 \\ \star 
		\end{bmatrix}, \begin{bmatrix}
		\mathcal{L}^{\circlearrowright}_{\mathbb{R}^3}(\mathcal{G}) \\ \star 
		\end{bmatrix}\right\}\subseteq \mathrm{null}(\bar{\mathcal{R}}_{\mathcal{G},d}). $$
		Note that the distance rigidity does not explicitly depend on the orientation of the nodes when expressed as a point in $SE(3)$.  This accounts for the free $\star$ entry in the subspace $\bar{S}_d$ corresponding to the rotations.}  Thus, the set of trivial motions in $\mathbb{R}^3$ can be seen as the projection of $\bar{S}_d$ in $\mathbb{R}^3$.
	
	\textcolor{black}{Similarly, for an $\mathsf{SE}(3)$ \textcolor{black}{bearing} framework one can define the subspace \cite{bearingSE3}
		$$\mathcal{S}_b \coloneqq \mathrm{span}\left\{ \begin{bmatrix}
		1_N \otimes I_3 \\ 0_{3N \times 3}, 
		\end{bmatrix}, \begin{bmatrix}
		p_\mathcal{G} \\ 0_{3N}, 
		\end{bmatrix}, \mathcal{L}^{\circlearrowright}_{\mathsf{SE}(3)}(\mathcal{G})\right\},$$
		where the vector $[p_\mathcal{G}^T \, 0_{3N}^T]^T$ represents a scaling of the framework. The space $\mathcal{L}^{\circlearrowright}_{\mathsf{SE}(3)}(\mathcal{G})$ is the rotational subspace induced by $\mathcal{G}$, in $\mathsf{SE}(3)$,
		\textcolor{black}{	\begin{equation} \label{eq:rot subspace SE3}
		\mathcal{L}^{\circlearrowright}_{\mathsf{SE}(3)}(\mathcal{G})=\mathrm{span}{\left\{\colvec{
				\left(  I_N \otimes S(\mathbf{e}_{h}) \right) p_{\mathcal{G}} \\
				1_N\otimes \mathbf{e}_h \\
			},{h=1,2,3} \right\} }. 
		\end{equation}}
		It is also known that $\mathcal{S}_b \subseteq \mathrm{null}(\mathcal{R}_{\mathcal{G},b})$. Thus $\mathcal{S}_b$ describes the trivial motions of an $\mathsf{SE}(3)$ \textcolor{black}{bearing} framework \cite{bearingSE3}.
	}
	%
	%
	\textcolor{black}{
		The above discussion 
		leads to the following result. 
		\begin{proposition}\label{prop.trivialmotions}
			The trivial motions of a D\&B framework are characterized by the set
			$$\mathcal{S}_{db} \coloneqq \bar{\mathcal{S}}_d \cap \mathcal{S}_b = \mathrm{span}\left\{ \begin{bmatrix}1_N \otimes I_3\\0_{3N\times 3} \end{bmatrix} , \mathcal{L}^{\circlearrowright}_{\mathsf{SE}(3)}(\mathcal{G}) \right\}.$$
			Furthermore, it follows that $\mathcal{S}_{db} \subseteq \mathrm{null}({\widetilde{\mathcal{R}}_{\mathcal{G}}})$.
		\end{proposition}
	}
	\textcolor{black}{Having characterized the trivial motions, it now follows from Definition \ref{def:inf rig} that for infinitesimal rigidity, we require that $\mathrm{null}({\widetilde{\mathcal{R}}_{\mathcal{G}}}) = \mathcal{S}_{db}$.  This is summarized as follows.
		\begin{proposition} \label{prop:ranks}
			The framework $(\mathcal{G},p_\mathcal{G},R_\mathcal{G})$ is D\&B infinitesimally rigid in $\mathsf{SE}(3)$ if and only if
			\begin{align}
			\textup{null}(\widetilde{\mathcal{R}}_\mathcal{G}) &=   \textup{null}(\bar{\mathcal{R}}_{\mathcal{G},d})\cap\textup{null}(\mathcal{R}_{\mathcal{G},b}) \notag \\ 
			&= \textup{span}\left\{ \begin{bmatrix}
			1_N \otimes I_3 \\
			0_{3N\times 3}, 
			\end{bmatrix}, \mathcal{L}^{\circlearrowright}_{\mathsf{SE}(3)}(\mathcal{G}) \right\}=\mathcal{S}_{db}.
			\end{align}
			Equivalently, the D\&B framework is infinitesimally rigid in $\mathsf{SE}(3)$ if and only if
			\begin{align}
			\textup{rank}(\widetilde{\mathcal{R}}_\mathcal{G}) &= \textup{dim}(\widetilde{\mathcal{R}}_\mathcal{G}) - \textup{dim}(\textup{null}(\widetilde{\mathcal{R}}_\mathcal{G})) = 6N - 6.
			\end{align}
		\end{proposition}
	}
	{Hence, all the motions produced by the nullspace of $\widetilde{\mathcal{R}}_\mathcal{G}$ for an infinitesimally rigid framework must correspond to trivial motions, i.e., translations and coordinated rotations.} 
	{Moreover, given \eqref{eq:R_G tilde}, it follows that $(\mathcal{G},p_\mathcal{G},R_\mathcal{G})$ is D\&B infinitesimally rigid in $\mathsf{SE}(3)$ if and only if
		\begin{align} \label{eq:R_G nullspace}
		\textup{null}(\mathcal{R}_\mathcal{G}) = \{x=P_Ry \in \mathsf{SE}(3)^N : y \in \textup{null}(\widetilde{\mathcal{R}}_\mathcal{G}) \},
		\end{align}
		i.e., {the nullspace of $\mathcal{R}_\mathcal{G}$ consists of} the vectors of $\textup{null}(\widetilde{\mathcal{R}}_\mathcal{G})$ whose elements are permutated by $P_R$. }
	
	{It is worth noting that the aforementioned results are not valid if the rigidity matrix loses rank, i.e., $\textup{rank}(\mathcal{R}_\mathcal{G}) < \max\{\textup{rank}(\mathcal{R}_\mathcal{G}(x)), x\in\mathsf{SE}(3)\}$. These are degenerate cases that correspond, for example, to when all agents are aligned along a direction $\mathbf{v} \in \mathsf{S}^2$.  {In particular, frameworks with $N=2$ nodes are also degenerate by this definition. } For more discussion on these cases, the reader is referred to \cite{Michieletto2021}. 
	
	{As a last remark, we observe that frameworks over the complete graph, $(\mathcal{K}_N, p_{\mathcal{K}_N},R_{\mathcal{K}_N})$, are (except for the degenerate configurations), infinitesimally rigid.  That is, $\textup{rank}(\widetilde{\mathcal{R}}_{\mathcal{K}_N}) =6N - 6$. {This result {follows from} the literature on distance- and $\mathsf{SE}(3)$-rigidity theory \cite{asimow1978rigidity, bearingSE3}}. This leads to the following corollary.
		\begin{corollary} \label{cor:R_g_tilde null}
			Consider the D\&B frameworks $(\mathcal{G},p_\mathcal{G},R_\mathcal{G})$ and $(\mathcal{K}_N,p_\mathcal{G},R_\mathcal{G})$ for nondegenrate configurations $(p_\mathcal{G},R_{\mathcal{G}})$.  Then $(\mathcal{G},p_\mathcal{G},R_\mathcal{G})$ is D\&B infinitesimally rigid if and only if
			$ \textup{rank}(\widetilde{\mathcal{R}}_{\mathcal{G}})= \textup{rank}(\widetilde{\mathcal{R}}_{\mathcal{K}_N}) = 6N-6.$
		\end{corollary}
	}
	
	{In the next section, we use the aforementioned results to link the D\&B rigidity matrix of a complete graph to the internal forces from \eqref{eq:interaction forces}}. 

	\section{Main Results} \label{sec:Main results}
	
    { 
    In cooperative manipulation schemes, the most energy-efficient way of transporting an object is to exploit the full potential of the cooperating robotic agents, i.e., each agent does not exert less effort at the expense of other agents, which might then potentially exert more effort than necessary. For instance, consider a rigid cooperative manipulation scheme, with 
	only one agent (a leader) working towards bringing the object to a desired location, whereas the other agents have zero inputs. Since the grasps are rigid, if the leader  {is equipped with sufficiently powerful actuators}, it will achieve the task by ``dragging" the rest of the agents, compensating for their dynamics, and creating non-negligible internal forces. In such cases, when the cooperative manipulation system is rigid (i.e., the grasps are considered to be rigid), the optimal strategy of transporting an object  is achieved by regulating the internal forces to zero. 	Therefore, from a control perspective, the goal of a rigid cooperative manipulation system is to design a control protocol that achieves a desired cooperative manipulation task, while guaranteeing that the internal forces {remain} zero. }
	
	{This section provides the main results of this work. 
	We first give, in Section \ref{sec:forces+rigid}, a closed-form expression for the internal forces of the {coupled object-agents system}, by connecting them with the D\&B rigidity matrix introduced in Section \ref{sec:Rigidity}. Next, we use these results in Section \ref{sec:manip internal} to provide a novel relation between the arising interaction and internal forces; we further give conditions on the agent force distribution for cooperative manipulation free from internal forces. }
    
	 
	\subsection{Internal Forces Based on the D\&B Rigidity Matrix} \label{sec:forces+rigid}
	
	{In this section we provide a closed-form expression for the internal forces of the coupled object-agents system and link them to the D\&B rigidity matrix notion introduced in Section \ref{sec:Rigidity}. In particular, we consider that the robotic agents form a graph {that will be defined in the sequel}}. {Note that, due to the rigidity of the grasping points, the forces exerted by an agent influence, not only the object, but all the other agents as well. Hence, since there exists interaction among all the pairs of agents 
	we model their connection as a complete graph, as {explicitly} described below}. Moreover, as will be clarified later, the rigidity matrix of this graph encodes the constraints among the agents, imposed by the rigidity of the grasping points, and plays an important role in the expression of the internal forces. }
	
	{Let the robotic agents form a  framework $(\mathcal{G},p_\mathcal{G},R_\mathcal{G})$ in $\mathsf{SE}(3)$, where $\mathcal{G}\coloneqq (\mathcal{N},\mathcal{E})$ is the complete graph, i.e., $\mathcal{E} = \{(i,j) \in \mathcal{N}^2 : i \neq j \}$, and $p_\mathcal{G} \coloneqq [p_1^\top,\dots,p_N^\top]^\top$, $R_\mathcal{G} \coloneqq (R_1,\dots,R_N)$. Consider also the undirected part $\mathcal{E}_u = \{(i,j)\in\mathcal{E} : i<j\}$ of $\mathcal{E}$, as also described in Section \ref{sec:Rigidity}. Since the graph is complete, we conclude that $|\mathcal{E}| = N(N-1)$ and $|\mathcal{E}_u| = \frac{N(N-1)}{2}$.} 
	
	{
		{Consider} now the rigidity functions
		$\gamma_{e,d}:\mathbb{R}^3\times\mathbb{R}^3\to\mathbb{R}_{\geq 0}$, $\forall e \in \bar{\mathcal{E}}_u$ and $\gamma_{e,b}:\mathsf{SE}(3)^2\to\mathsf{S}^2$, $\forall e \in {\mathcal{E}}$, as given in \eqref{eq:gamma 1}, as well as the stack vector $\gamma_{{\mathcal{G}}}: \mathsf{SE}(3)^{{N}} \to \mathbb{R}^{\frac{{N}({N}-1)}{2}}\times\mathsf{S}^{2{N}({N}-1)}$ as given in \eqref{eq:gamma G}.
		The rigidity constraints of the framework are encoded in the constraint $\gamma_{{\mathcal{G}}} = \text{const.}$.  Since the rigidity of the framework stems from the rigidity of the grasping points, these constraints encode also the rigidity constraints of the object-agent cooperative manipulation.
		{By differentiating twice $\gamma_{{\mathcal{G}}}=\textup{const.}$}, one obtains  
				\begin{align}   \label{eq:A_int, b_int}
					 {\mathcal{R}_{{\mathcal{G}}}({x}) \dot{{v}} 
					=  - \dot{\mathcal{R}}_{{\mathcal{G}}}({x}) {v}}
				\end{align}
			where $\mathcal{R}_{{\mathcal{G}}}:\mathsf{SE}(3)^{{N}}\to \mathbb{R}^{\frac{7{N}({N}-1)}{2}\times(6{N})}$ is the rigidity matrix associated to ${\mathcal{G}}$ and has the form  \eqref{eq:rigidity matrix}.    
			{Note that  \eqref{eq:A_int, b_int} is derived from $\gamma_{{\mathcal{G}}}$, which corresponds to the distance and bearing constraints for a complete graph; hence, using Corollary \ref{cor:R_g_tilde null}, $\gamma_{{\mathcal{G}}}$ encodes rigid body motions (coordinated translations and rotations of the system). Therefore, by assuming that 
			the agents satisfy the constraint  $\dot{\gamma}_{{\mathcal{G}}} = \mathcal{R}_{{\mathcal{G}}}{v} = 0$ initially\footnote{{Otherwise, the constraint $ \mathcal{R}_{{\mathcal{G}}}{v} = 0$ can be appended in \eqref{eq:A_int, b_int}.}}, we conclude that the motion of the cooperative object-agents manipulation system that is enforced by  \eqref{eq:A_int, b_int},  corresponds to rigid body motions (coordinated translations and rotations of the system). Hence, since ${\mathcal{G}}$ is complete, the analysis of Section \ref{sec:Rigidity} dictates 
		that these motions are the infinitesimal motions of the framework and are the ones produced by the nullspace of  $\mathcal{R}_{{\mathcal{G}}}({x} )$. We note that the rigid body motions \textit{can} be produced by the nullspace of the rigidity matrix of other graph topologies as well (except for the complete one). Nevertheless, as explained above, the complete graph topology draws motivation from the physics of the cooperative manipulation system, where all agents indeed influence the object as well as each other via their exerted forces.}
		}

	{{After giving the rigidity constraints in the cooperative manipulation system, we are now ready to derive the expressions for the internal forces, $h_\textup{int}$, in terms of the aforementioned rigidity matrix. We follow the same methodology as in \cite{erhart2015internal}. Since we are concerned with the internal forces, consider, without loss of generality, that $h_{\scr O} = h_\textup{m} = 0 \Leftrightarrow h =  h_\textup{int}$, i.e., the agents produce only internal forces, without inducing object acceleration. Then, the agent dynamics are \begin{align} \label{eq:agent dynamics only internal force}
        {M(x)\dot{v} + C(x,\dot{x})v + g(x)  = u - h_{\text{int}} }
    \end{align}
    }
	{We use Gauss' principle \cite{Udwadia92NewPerspective,Udwadia93Constrained} {(see Lemma \ref{lem:gauss})} to derive a closed form expression for $h_\textup{int}$. 
	Let the \textit{unconstrained} system of the robotic agents be
	{${M}({x})\alpha({x},\dot{{x}})
				\coloneqq {u} - {C}({x},{\dot{x}})
				{v} - {g}({x}),$}
	where $\alpha$ is the unconstrained acceleration, i.e., the acceleration the system would have if the agents did not grasp the object. According to Gauss's principle \cite{Udwadia92NewPerspective}, the actual acceleration ${\dot{v}}$ of the system is the closest one to $\alpha$, while satisfying the rigidity constraints. More rigorously, {and as stated in Lemma \ref{lem:gauss},} $\dot{{v}}$ is the solution of the constrained minimization problem 
		{\begin{align*}
					&\min \ \ [\dot{{v}} - \alpha({x},\dot{{x}}) ]^\top {M}({x}) [ \dot{{v}} - \alpha({x},\dot{{x}}) ] \\ 		
					&\hspace{1mm} \textup{s.t.} \ \ \ \mathcal{R}_{{\mathcal{G}}}({x}) \dot{{v}} 
					=  - \dot{\mathcal{R}}_{{\mathcal{G}}}({x}) {v}. 
			\end{align*}}
			The solution to this problem is obtained by using the 
			Karush-Kuhn-Tucker conditions \cite{boyd2004convex} and has a closed-form expression. It can be shown that it satisfies 
				\begin{align*}
					{{M}\dot{{v}} = {{M}} \alpha -  \mathcal{R}_{{\mathcal{G}}}^\top ( \mathcal{R}_{{\mathcal{G}}} {M}^{-1} \mathcal{R}_{{\mathcal{G}}}^\top )^\dagger (\dot{\mathcal{R}}_{{\mathcal{G}}} {v} + \mathcal{R}_{{\mathcal{G}}} \alpha ),}
			\end{align*}
			which 
			is {consistent}  with the one in \cite{Udwadia92NewPerspective},
			\begin{equation*}
					{{M}\dot{{v}} = {{M}} \alpha - {M}^{\frac{1}{2}} ( \mathcal{R}_{{\mathcal{G}}}  {M}^{-\frac{1}{2}}) ^\dagger ( \dot{\mathcal{R}}_{{\mathcal{G}}} {v}  + \mathcal{R}_{{\mathcal{G}}}\alpha ),}
			\end{equation*} 
			since it holds that $\mathcal{R}_{{\mathcal{G}}}^\top ( \mathcal{R}_{{\mathcal{G}}} {M}^{-1} \mathcal{R}_{{\mathcal{G}}}^\top )^\dagger = {M}^{\frac{1}{2}} ( \mathcal{R}_{{\mathcal{G}}}  {M}^{-\frac{1}{2}}) ^\dagger$. Indeed, {according to Property 1) of Section \ref{sec:preliminaries}}, it holds that $H^\dagger = H^\top (H H^\top)^\dagger$, for any $H\in\mathbb{R}^{x\times y}$. Then the aforementioned equality is obtained by setting $H = \mathcal{R}_{{\mathcal{G}}} {M}^{-\frac{1}{2}}$.}} 
		
		{Therefore, the internal forces have the form
		\begin{subequations}	\label{eq:internal forces 1}
			{
				\begin{align} 
					h_\textup{int} &= \mathcal{R}_{\mathcal{G}}^\top  (\mathcal{R}_{\mathcal{G}} M^{-1} \mathcal{R}_{\mathcal{G}}^\top )^\dagger ( \dot{\mathcal{R}}_{\mathcal{G}} v +    \mathcal{R}_{\mathcal{G}} \alpha) \label{eq:internal forces 1_1} \\
					&= M^{\frac{1}{2}} ( \mathcal{R}_{\mathcal{G}}   M^{-\frac{1}{2}} )^\dagger ( \dot{\mathcal{R}}_{\mathcal{G}} v + \mathcal{R}_{\mathcal{G}} \alpha ) \label{eq:internal forces 1_2}
			\end{align} }
		\end{subequations}
		and one concludes that when the unconstrained motion of the system does not satisfy the constraints (i.e., when $\dot{\mathcal{R}}_{\mathcal{G}} v \neq - {\mathcal{R}}_{\mathcal{G}} \alpha$), then the actual accelerations of the system are modified in a manner directly proportional to the extent to which these constraints are violated. Moreover, it is evident from the aforementioned expression that the internal forces depend, not only on the relative distances $p_i - p_j$, but also on the closed loop dynamics and the inertia of the unconstrained system (see the dependence on $\alpha_\textup{int}$ and $B$). Therefore, given a desired force $h_{\text{d}}$ to be applied to the object, an internal force-free distribution to agent forces $h_{i,\text{d}}$ at the grasping points cannot be independent of the system dynamics. {We stress that the derived expression concerns the internal forces produced exclusively by the redundancy of the multi-robot system (excluding, for instance, potential internal forces needed to keep the object from falling due to gravity, which would also arise in a single manipulation task). }
		{By following a similar procedure and including the object in the multi-agent graph, one can arrive to similar results for the interaction forces $h$ as well.}
		

		Note that, as dictated in Section \ref{sec:Rigidity}, the rigidity matrix $\mathcal{R}_\mathcal{G}$ is not unique, since different choices of $\gamma_\mathcal{G}$ that encode the rigidity constraints can be made.  Hence, one might think that different expressions of $\mathcal{R}_\mathcal{G}$ will result in different rigidity constraints of the form \eqref{eq:A_int, b_int} and hence different internal forces - which is unreasonable. {Therefore, in order to show the  consistency of \eqref{eq:internal forces 1}, we prove next in Proposition \ref{corol:different R same h_int} that this is not the case, by using the fact that 
			all different expressions of the rigidity matrix $\mathcal{R}_\mathcal{G}$ have the same nullspace (the coordinated translations and rotations of the framework).}
		
		\begin{proposition} \label{corol:different R same h_int}
			Let $\mathcal{R}_{\mathcal{G},1}$ and $\mathcal{R}_{\mathcal{G},2}$ such that $\textup{null}(\mathcal{R}_{\mathcal{G},1})=\textup{null}(\mathcal{R}_{\mathcal{G},2})$ and let
			{
				\begin{align*}
					h_{\textup{int},i} \coloneqq& M^{\frac{1}{2}} ( \mathcal{R}_{\mathcal{G},i} M^{-\frac{1}{2}} )^\dagger ( \dot{\mathcal{R}}_{\mathcal{G},i}{v} + {\mathcal{R}}_{\mathcal{G},i} \alpha_{\textup{int}} ), 
					\ \  \forall i\in\{1,2\}.
				\end{align*}    	
				Then $h_{\textup{int},1} = h_{\textup{int},2}$.  }
		\end{proposition}    
	{		\begin{IEEEproof} 
    	The poses and velocities in the terms {$\dot{\mathcal{R}}_{\mathcal{G},i}{v}$} are the actual ones resulting from the coupled system dynamics and hence they respect the rigidity constraints imposed by {$R_{\mathcal{G},i} \dot{v} = - \dot{\mathcal{R}}_{\mathcal{G},i}{v}$, for $i\in\{1,2\}$}. Therefore, exploiting the positive definiteness of $M$, we need to prove that {$( \mathcal{R}_{\mathcal{G},1} M^{-\frac{1}{2}})^\dagger \mathcal{R}_{\mathcal{G},1} = ( \mathcal{R}_{\mathcal{G},2}  M^{-\frac{1}{2}} )^\dagger \mathcal{R}_{\mathcal{G},2}$}. 
    	{According to property 4) of Sec. \ref{sec:preliminaries}}, since $\mathcal{R}_{\mathcal{G},1}$ and $\mathcal{R}_{\mathcal{G},2}$ have the same nullspace, they are left equivalent matrices and there exists an invertible matrix $P$ such that $\mathcal{R}_{\mathcal{G},1} = P \mathcal{R}_{\mathcal{G},2}$. Hence, {by further using property 2) of  Sec. \ref{sec:preliminaries}}, it holds that
			{\begin{align*}
					& ( \mathcal{R}_{\mathcal{G},2} M^{-\frac{1}{2}} )^\dagger \mathcal{R}_{\mathcal{G},2}  -
					( \mathcal{R}_{\mathcal{G},1} M^{-\frac{1}{2}} )^\dagger \mathcal{R}_{\mathcal{G},1} = \\ 
					&\bigg[ ( \mathcal{R}_{\mathcal{G},2} M^{-\frac{1}{2}} )^\dagger \mathcal{R}_{\mathcal{G},2} M^{-\frac{1}{2}} - 
					( P \mathcal{R}_{\mathcal{G},2} M^{-\frac{1}{2}} )^\dagger P \mathcal{R}_{\mathcal{G},2} M^{-\frac{1}{2}}  \bigg] M^{\frac{1}{2}}, 
				\end{align*}
			}
			which is {equal to} $0$. 
		\end{IEEEproof} }

        {Additionally, by following its proof, one can conclude that Proposition \ref{corol:different R same h_int} can be extended to account for different graphs $\mathcal{G}$ and $\mathcal{G}'$ satisfying \eqref{eq:A_int, b_int}.    }
		
		{Next, we note that \eqref{eq:internal forces 1} leads to the following Lemma:}
		
		\begin{lemma} \label{coroll:A and b}
			The cooperative manipulation system is free of internal forces, i.e., {$h_{\textup{int}} =  0_{6N}$}, if and only if 
			\begin{equation*}
				\dot{\mathcal{R}}_\mathcal{G}v  + \mathcal{R}_\mathcal{G} M^{-1}(u - C\dot{v} -g) \in \textup{null}(\mathcal{R}_\mathcal{G}^\top)
			\end{equation*}
		\end{lemma}
		{
	    \begin{IEEEproof}
	    In view of \eqref{eq:internal forces 1}, $\dot{\mathcal{R}}_\mathcal{G} v + \mathcal{R}_\mathcal{G} \alpha_{\textup{int}} = \dot{\mathcal{R}}_\mathcal{G}v  + \mathcal{R}_\mathcal{G} M^{-1}(u - C\dot{v} -g)$ must belong to $\textup{null}(M^{\frac{1}{2}} (\mathcal{R}_\mathcal{G} M^{-\frac{1}{2}})^\dagger)$ in order to avoid internal forces. The latter, however, is identical to  $\textup{null}(\mathcal{R}_\mathcal{G}^\top)$, since it holds that $\textup{null}(\mathcal{R}_\mathcal{G} M^{-1/2})^\dagger = \textup{null}(M^{-\frac{1}{2}}\mathcal{R}_\mathcal{G}^\top)$ and $M$ is  positive definite. 
	    \end{IEEEproof}}
	    {As mentioned before, the most energy-efficient way of transporting an object in a cooperative manipulation scheme is {by minimizing} the arising internal forces. }	
		{In the next section, we derive a new relation between the interaction and internal forces as well as novel sufficient and necessary conditions on the agent force  distribution  for  the  provable  regulation  of  the  internal forces to zero, according to \eqref{eq:internal forces 1}. We further show its application in  a  standard  inverse-dynamics  control  law  that  guarantees trajectory tracking by the object’s center of mass. }

	\subsection{Cooperative Manipulation via Internal Force Regulation} \label{sec:manip internal}
	
	{In this section, we use the results of Section \ref{sec:forces+rigid} to derive a new relation between the interaction and internal forces $h$ and $h_{\text{int}}$, respectively.}} Moreover,
	we	derive novel sufficient and necessary conditions on the agent force distribution for the provable regulation of the internal forces to zero, according to \eqref{eq:internal forces 1}, and we show its application in a standard inverse-dynamics control law that guarantees trajectory tracking of the object's center of mass. This is based on the following {main theorem}, which links the complete agent graph rigidity matrix $\mathcal{R}_\mathcal{G}$ to the grasp matrix $G$: }
	
	\begin{theorem} \label{th:null G range R_T}
		Let $N$ robotic agents, with configuration $x = (p,R)\in\mathsf{SE}(3)^N$, rigidly grasping an object and associated with a grasp matrix $G(x)$, as in \eqref{eq:grasp matrix def.}. Let also the agents be 		
		modeled by a framework on the complete graph $(\mathcal{K}_N, p_{\mathcal{K}_N}, R_{\mathcal{K}_N}) = (\mathcal{K}_N, p, R)$ in $\mathsf{SE}(3)$, which is associated with a rigidity matrix $\mathcal{R}_{\mathcal{K}_N}$. 
		 Let also $x$ be such that $\textup{rank}(\mathcal{R}_{\mathcal{K}_N}(x)) = {\max_{y \in \mathsf{SE}(3)^N}\{ \textup{rank}(\mathcal{R}_{\mathcal{K}_N}(y))  \}}$.
		Then 
		\begin{equation} \label{eq:null_G = range R}
			\textup{null}(G(x)) = \textup{range}(\mathcal{R}_{\mathcal{K}_N}(x)^\top).
		\end{equation}
	\end{theorem}
	{
	\begin{IEEEproof}
			Since $\mathcal{R}_{\mathcal{K}_N}$ {corresponds to} the complete graph and $\textup{rank}(\mathcal{R}_{\mathcal{K}_N}(x)) = \max_{y \in \mathsf{SE}(3)^N}\{ \textup{rank}(\mathcal{R}_{\mathcal{K}_N}(y))  \}$, the framework $(\mathcal{K}_N, p, R)$ is infinitesimally rigid. 
			Hence, the nullspace of $\mathcal{R}_{\mathcal{K}_N}$ consists only of the infinitesimal motions of the framework, i.e., coordinated translations and rotations, as defined in Prop. \ref{prop.trivialmotions}. 		
			In particular, in view of \eqref{eq:R_G nullspace}, {Prop. \ref{prop:ranks}}, and \eqref{eq:rot subspace SE3}, one concludes that $\textup{null}(\mathcal{R}_{\mathcal{K}_N})$ is the linear span of $1_N\otimes \begin{bmatrix}
				I_3 \\ 0_{3 \times 3}
			\end{bmatrix}$ and the vector space
			$[\chi_1^\top, \dots, \chi_N^\top]^\top \in \mathsf{SE}(3)^N$, with $\chi_i\coloneqq [\chi_{i,p}^\top, \chi_{i,R}^\top]^\top\in\mathsf{SE}(3)$, satisfying
			\begin{subequations} \label{eq:rot subsp}
				\begin{align}
					& \chi_{i,p} - \chi_{j,p} = -S(p_i - p_j) \chi_{i,R} \\
					& \chi_{i,R} = \chi_{j,R},
				\end{align}
			\end{subequations}
			where $p_i \coloneqq p_{\mathcal{K}_N}(i)$, $p_j \coloneqq p_{\mathcal{K}_N}(j)$, $\forall i,j \in\mathcal{N}$, with $i\neq j$. In view of \eqref{eq:grasp matrix velocities}, one obtains $v = G(x)^\top v_{\scr O}$. 
			Note that the first $3$ columns of $G^\top$ form the space $1_N\otimes \begin{bmatrix}
				I_3 \\ 0_{3 \times 3}
			\end{bmatrix}$ whereas its last $3$ columns span the aforementioned rotation vector space. Indeed, for any  $\dot{p}_{\scr O}$, $\omega_{\scr O} \in \mathbb{R}^6$ the range of these columns is
			\begin{align*}
				\begin{bmatrix}
					-\dot{p}_{\scr O}^\top S(p_{1\scr O})^\top, 
					\omega_{\scr O}^\top,\dots,	
					-\dot{p}_{\scr O}^\top S(p_{N \scr O})^\top,
					\omega_{\scr O}^\top
				\end{bmatrix}^\top,
			\end{align*}
			for which it is straightforward to verify that \eqref{eq:rot subsp} holds.	Hence, $\textup{null}(\mathcal{R}_{\mathcal{K}_N}) = \textup{range}(G^\top)$, which implies \eqref{eq:null_G = range R}.
		\end{IEEEproof} }
		
	{We note that, in degenerate cases (i.e., when $\mathcal{R}_\mathcal{G}$ loses rank, as explained in Sec. \ref{sec:Rigidity}), $\textup{null}(\mathcal{R}_\mathcal{G})$ contains more motions than the trivial coordinated translations and rotations, i.e., $\textup{range}(G^\top) \subset  \textup{null}(\mathcal{R}_\mathcal{G})$. Therefore, \eqref{eq:null_G = range R} is replaced by $\textup{range}(\mathcal{R}_\mathcal{G}^\top) \subset \textup{null}(G)$.}   
	
	{Since the internal forces belong to $\text{null}(G)$}, one concludes that they are comprised of all the vectors $z$ for which there exists a $y$ such that $z = \mathcal{R}_\mathcal{G}^\top y$. This can also be verified by inspecting \eqref{eq:internal forces 1_2}; one can prove that $\textup{range}( M^{\frac{1}{2}} (\mathcal{R}_\mathcal{G} M^{-\frac{1}{2}})^\dagger ) = \textup{range}(\mathcal{R}_\mathcal{G}^\top)$. The aforementioned result provides significant insight regarding the  control of the motion of the coupled cooperative manipulation system. In particular,  
	by using \eqref{eq:internal forces 1_2} and Th. \ref{th:null G range R_T}, we provide next {new conditions on the agent force distribution for provable avoidance of internal forces. We first derive a novel relation between the agent forces $h$ and the internal forces $h_\text{int}$. }
	
	{
		In many related works, $h$ is decomposed as 
		\begin{equation} \label{eq:force decomp}
		h =  G^\ast G h + (I - G^\ast G)h,
		\end{equation}
		where $G^\ast$ is a right inverse of $G$. The term $G^\ast G h$ is a projection of $h$ on the range space of $G^\top$, whereas the term $(I - G^\ast G)h$ is a projection of $h$ on the null space of $G$. A common choice is the Moore-Penrose inverse $G^\ast = G^\dagger=G^\top (GG^\top)^{-1}$. This specific choice yields the vector $G^\ast G h=G^\dagger G h \in \textup{range}(G^\top)$ that is closest to $h$, i.e., $\|h - G^\dagger G h\| \leq \|h - y\|$, $\forall y\in\textup{range}(G^\top)$. However, as the next theorem states, if the second term of \eqref{eq:force decomp} must equal $h_\text{int}$, as defined in \eqref{eq:internal forces 1}, $G^\ast$ must equal $MG^\top(GMG^\top)^{-1}$.}
		
		{
		\begin{theorem} \label{th:internal forces}
			Consider $N$ robotic agents rigidly grasping an object 
			with coupled dynamics \eqref{eq:coupled dynamics}. Let $h \in \mathbb{R}^{6N}$ be the stacked vector of agent forces exerted at the grasping points. Then the agent forces $h$ and the internal forces $h_\textup{int}$ are related as:
			\begin{equation*}
			h_\textup{int} =  (I_{6N} - MG^\top(GMG^\top)^{-1}G) h.
			\end{equation*}
		\end{theorem}
	{
	    In order to prove Th. \ref{th:internal forces}, we first need the following result.
			\begin{proposition} \label{prop:forces 2}
				Consider the grasp and rigidity matrices $G$, $\mathcal{R}_\mathcal{G}$, of \eqref{eq:grasp matrix def.}, \eqref{eq:rigidity matrix}, respectively. Then it holds that 
				\begin{equation} \label{eq:internal forces wrt to forces}
					MG^\top (G M G^\top)^{-1} G + M^\frac{1}{2}(\mathcal{R}_\mathcal{G}M^{-\frac{1}{2}})^\dagger \mathcal{R}_\mathcal{G} M^{-1} = I.
				\end{equation}		
			\end{proposition}
			\begin{IEEEproof}
				Let $A \coloneqq  \mathcal{R}_\mathcal{G} M^{-\frac{1}{2}} $ and $B \coloneqq G M^{\frac{1}{2}}$. Then $\textup{range}(A^\top) = \textup{null}(B)$. Indeed, according to Th. \ref{th:null G range R_T}, it holds that if $z = \mathcal{R}^\top_\mathcal{G} y$, for some $y\in\mathbb{R}^6$, then $G z =0_6$. By multiplying by $M^{-\frac{1}{2}}$, we obtain $M^{-\frac{1}{2}} z = M^{-\frac{1}{2}} \mathcal{R}^\top_\mathcal{G}y$, which implies that $\hat{z}\coloneqq M^{-\frac{1}{2}}z \in \textup{range}( (\mathcal{R}_\mathcal{G}M^{\frac{1}{2}})^\top)$. It also holds that $B\hat{z} =  GM^{\frac{1}{2}}\hat{z} = Gz = 0_6$, and hence $\hat{z}\in\textup{null}(B)$. Therefore, {by using properties 1) and 3) of Sec. \ref{sec:preliminaries}} 
				and the fact that $GMG^\top$ is invertible, we conclude that 
				\begin{align*}
					&(G M^{\frac{1}{2}})^\dagger GM^{\frac{1}{2}} + (\mathcal{R}_\mathcal{G}M^{-\frac{1}{2}})^\dagger\mathcal{R}_\mathcal{G}M^{-\frac{1}{2}} = I \Leftrightarrow \\
					&M^\frac{1}{2}G^\top(G M G^\top)^\dagger GM^\frac{1}{2} + (\mathcal{R}_\mathcal{G}M^{-\frac{1}{2}})^\dagger\mathcal{R}_\mathcal{G}M^{-\frac{1}{2}} = I,
				\end{align*}
				and by left and right multiplication by $M^{\frac{1}{2}}$ and $M^{-\frac{1}{2}}$, respectively, the result follows.
			\end{IEEEproof}
		{Moreover, for the proof of Th. \ref{th:internal forces}, we need the following expression, which is derived from  \eqref{eq:grasp matrix velocities}, \eqref{eq:manipulator dynamics_task space vector_form}, \eqref{eq:object dynamics}, and { \eqref{eq:grasp matrix}.}
		\begin{align} \label{eq:h 1}
			h =& (M^{-1} + G^\top M_{\scr O}^{-1} G )^{-1} [ M^{-1}(u - g - C v) - \dot{G}^\top v_{\scr O}  \notag\\
			&+ G^\top M_{\scr O}^{-1}(C_{\scr O}v_{\scr O} + g_{\scr O}) ].
		\end{align}}
			\begin{IEEEproof}[Proof of Theorem \ref{th:internal forces}]
				We first show that
				\begin{align*}		
					&	[I - MG^\top(GMG^\top)^{-1}G](M^{-1} + G^\top M_{\scr O}^{-1}G)^{-1} = \\ &\hspace{5cm} M^\frac{1}{2}(\mathcal{R}_\mathcal{G}M^{-\frac{1}{2}})^\dagger \mathcal{R}_\mathcal{G}.
				\end{align*}
				Indeed, since $(M^{-1} + G^\top M_{\scr O}^{-1}G)^{-1}$ has full rank, it suffices to show that
				\begin{align*}
					&I - MG^\top(GMG^\top)^{-1}G = \\
					&\hspace{10mm} M^\frac{1}{2}(\mathcal{R}_\mathcal{G}M^{-\frac{1}{2}})^\dagger \mathcal{R}_\mathcal{G} (M^{-1} + G^\top M_{\scr O}^{-1}G), 
				\end{align*}
				which can be concluded from the fact that $\mathcal{R}_\mathcal{G} G^\top = 0$ (due to Th. \ref{th:null G range R_T}) and Prop. \ref{prop:forces 2}.
				Therefore, in view of \eqref{eq:h 1}, it holds that 
				\small
				\begin{align*}
					&(I - MG^\top(GMG^\top)^{-1}G)h =\\
					& [I - MG^\top(GMG^\top)^{-1}G](M^{-1} + G^\top M_{\scr O}^{-1} G)^{-1}
					[- \dot{G}^\top v_{\scr O} \\&
					+M^{-1}(u - g - C v) + G^\top M_{\scr O}^{-1}(C_{\scr O}v_{\scr O} + g_{\scr O}) ] = \\
					&M^\frac{1}{2} (\mathcal{R}_\mathcal{G} M^{-\frac{1}{2}})^\dagger\mathcal{R}_\mathcal{G} [M^{-1}(u - g - C v) + G^\top M_{\scr O}^{-1}(C_{\scr O}v_{\scr O} + g_{\scr O}) \\
					&- \dot{G}^\top v_{\scr O}]
				\end{align*}
				\normalsize
				which, in view of the facts that $\mathcal{R}_\mathcal{G} G^\top = 0$, and hence $-\mathcal{R}_\mathcal{G} \dot{G}^\top = \dot{\mathcal{R}}_\mathcal{G} G^\top$, as well as $G^\top v_{\scr O} = v$, becomes
				\begin{align*}
					M^\frac{1}{2}(\mathcal{R}_\mathcal{G} M^{-\frac{1}{2}})^\dagger[\dot{\mathcal{R}}_\mathcal{G}v + \mathcal{R}_\mathcal{G}M^{-1}(u - g - C v)] = h_\text{int}.
				\end{align*}
			\end{IEEEproof}  }

	{
		Based on Th. \ref{th:internal forces}, we provide next new results on the internal force-free (optimal) distribution of a force to the agents.
		}
		{
	\begin{theorem} \label{th:optimal internal force distribution}
			Consider $N$ robotic agents rigidly grasping an object 
			with coupled dynamics \eqref{eq:coupled dynamics}. Let a desired force to be applied to the object $h_{\scr O,\textup{d}} \in \mathbb{R}^6$, which is distributed to the agents' desired forces as $h_\textup{d} = G^\ast h_{\scr O,\textup{d}}$, and where $G^\ast$ is a right inverse of $G$, i.e., $GG^\ast = I_6$. Then it holds that 
		\begin{equation*}
			h_\textup{int} = 0 \Leftrightarrow G^\ast =   MG^\top(GMG^\top)^{-1}.
		\end{equation*}
	\end{theorem}}
	
    \begin{IEEEproof} 
				{
				{  {Firstly, it is easily verified that $G^\ast$ is a pseudo-inverse of $G$ \cite{albert72Pseudoinverse}}. Next,				according to Th. \ref{th:internal forces}, the derivation of $h_\text{d}$ that yields zero internal forces can be formulated as a quadratic minimization problem:			
					\begin{align}
						\text{QP}: \hspace{2mm }&\min_{h_\text{d}} \hspace{5mm} \|h_\textup{int}\|^2 =  h_\text{d}^\top H h_\textup{d} 	 				&\text{s.t. }  \hspace{3.5mm} Gh_\textup{d} = h_{\scr O,\textup{d}},
					\end{align}
					where $H \coloneqq (I_{6N} - MG^\top(GMG^\top)^{-1}G)^\top(I_{6N} - MG^\top(GMG^\top)^{-1}G)$. }
				{
					Note that the choice $G^\ast = MG^\top(GMG^\top)^{-1}h_{\scr O,\textup{d}}$ is a minimizer of QP, since $GG^\ast = I_6$, and $H G^\ast h_{\scr O,\textup{d}} = 0_{6N}$, and therefore sufficiency is proved. }}
				
				{In order to prove necessity, we prove next that $G^\ast$ is a strict minimizer, i.e., there is no other right inverse of $G$ that is a solution to of QP. Note first that $G\in\mathbb{R}^{6\times6N}$ has full row rank, which implies that the dimension of its nullspace is $6N-6$. Let $Z \coloneqq [z_1,\dots,z_{6N-6}] \in \mathbb{R}^{6N\times(6N-6)}$ be the matrix formed by the vectors $z_1,\dots,z_{6N-6}\in \mathbb{R}^{6N}$ that span the nullspace of $G$. It follows that $\textup{rank}(Z) = 6N-6$ and $GZ = 0_{6\times 6N-6}$. Let now the matrix $H' \coloneqq Z^\top H Z \in\mathbb{R}^{(6N-6)\times(6N-6)}$. Since $GZ = 0_{6\times {(6N-6)}} \Rightarrow Z^\top G^\top = 0_{(6N-6)\times6}$, it follows that $H' = Z^\top Z$. Hence, $\textup{rank}(H') = \textup{rank}(Z) = 6N-6$, which implies that $H'$ is positive definite. Therefore, according to  \cite[Th. $1.1$]{gould1985practical}, QP has a strong minimizer. 
				}		
			\end{IEEEproof} 
	
	{
		The aforementioned theorem provides novel necessary and sufficient conditions for provable minimization of internal forces in a cooperative manipulation scheme. As discussed before, this is crucial for achieving energy-optimal cooperative manipulation, where the agents do not have to ``waste" control input and hence energy resources that do not contribute to object motion. 
		Related works that focus on deriving internal force-free distributions $G^\ast$, e.g., \cite{walker1991analysis,chung2005analysis,williams1993virtual,erhart2015internal,donner2018physically}, are solely based on the inter-agent distances, neglecting the actual dynamics of the agents and the object. The expressions \eqref{eq:internal forces 1}, however, give new insight on the topic and suggest that the dynamic terms of the system play a significant role in the arising internal forces, as also indicated by Coroll. \ref{coroll:A and b}. This is further exploited by Th. \ref{th:optimal internal force distribution} to derive a right-inverse that depends on the inertia of the system. Note also that, as mentioned before and explained in \cite{erhart2015internal}, the internal forces depend on the acceleration of the robotic agents and hence the incorporation of $M$ in $G^\ast$ is something to be expected.
		}

	{
		The forces $h$, however, are not the actual control input of the robotic agents, and hence we cannot simply set $h = h_\textup{d} = MG^\top(GMG^\top)^{-1}Gh_{\scr O,\textup{d}}$ for a given $h_{\scr O,\textup{d}}$. Therefore, we design next a standard inverse-dynamics control algorithm controller that guarantees tracking of a desired trajectory by the object center of mass while provably achieving regulation of the internal forces to zero. 
		{Provable force regulation is also done in \cite{aghili2005unified}, requiring however the constraints matrix ($\mathcal{R}_\mathcal{G}$ in our case) to have positive singular values.}
	 }	
	 
	\subsection{Control Design}

	Let a desired position trajectory for the object center of mass be $p_\textup{d}:\mathbb{R}_{\geq 0}\to\mathbb{R}^3$, and $e_p \coloneqq p_{\scr O} - p_\textup{d}$. Let also a desired object orientation be expressed in terms of a desired rotation matrix $R_\textup{d}:\mathbb{R}_{\geq 0} \to \mathsf{SO}(3)$, with $\dot{R}_\textup{d} = S(\omega_\textup{d})R_\textup{d}$, where $\omega_\textup{d}:\mathbb{R}_{\geq 0} \to\mathbb{R}^3$ is the desired angular velocity. Then an orientation error metric is \cite{Automatica_formation_18} $e_{\scr O} \coloneqq \frac{1}{2}\textup{tr}\left(I_3 - R_\textup{d}^\top R_{\scr O} \right) \in [0,2]$,
	which, after differentiation and by using \eqref{eq:object dynamics 1} and properties of skew-symmetric matrices, becomes \cite{Automatica_formation_18}
	\begin{equation} \label{eq:e_O_dot rot mat}
	\dot{e}_{\scr O} = \frac{1}{2}e_R^\top R^\top_{\scr O}\left( \omega_{\scr O} - \omega_\textup{d} \right),
	\end{equation}
	where $e_R \coloneqq S^{-1}\left( R_\textup{d}^\top R_{\scr O} - R^\top_{\scr O} R_\textup{d} \right) \in\mathbb{R}^3$. The equilibrium $e_R = 0$ corresponds to $e_{\scr O} = 0$, implying $\textup{tr}(R_\textup{d}^\top R_{\scr O}) = 3$ and $R_{\scr O} = R_\textup{d}$, as well as to $e_{\scr O} = 2$ implying $\textup{tr}(R_\textup{d}^\top R_{\scr O}) = -1$ and $R_{\scr O} \neq R_\textup{d}$ \cite{Automatica_formation_18}.
	The second case represents an undesired equilibrium, where the desired and the actual orientation differ by $180$ degrees. This issue is caused by topological obstructions on $\mathsf{SO}(3)$ and it has been proven that no continuous controller can achieve \textit{global} stabilization \cite{Automatica_formation_18}. We design next a control protocol that guarantees internal force-free convergence of $e_p$, $e_{\scr O}$, while guaranteeing that $e_{\scr O}(t) < 2$, $\forall t\in\mathbb{R}_{\geq 0}$, 
		  provided that the right inverse $G^\ast = MG^\top(GMG^\top)^{-1}$ is used.
	
	{
		\begin{corollary} \label{corol:u design rot mat}
		Consider $N$ robotic agents rigidly grasping an object, as described in Section \ref{sec:Coop manip model}, with coupled dynamics \eqref{eq:coupled dynamics}. Let a desired trajectory be defined by $p_\textup{d}:\mathbb{R}_{\geq 0}\to\mathbb{R}^{3}$, $R_\textup{d}:\mathbb{R}_{\geq 0} \to \mathsf{SO}(3)$, {$\dot{p}_\textup{d}, \omega_\textup{d}\in\mathbb{R}^3$}, and assume that $e_{\scr O}(0) < 2$. 
		Consider the control law
		\begin{align} 
		u &= g + ( C G^\top + M \dot{G}^\top ) v_{\scr O} +  G^\ast\left( g_{\scr O} + C_{\scr O} v_{\scr O} \right) + \notag \\
		& (M G^\top + G^\ast M_{\scr O} )( \dot{v}_\textup{d} - K_d e_v - K_p e_x  ), \label{eq:u rot mat}
		\end{align}
		where $e_v \coloneqq v_{\scr O} - v_\textup{d}$, $v_\textup{d} \coloneqq [\dot{p}_\textup{d}^\top, \omega_\textup{d}^\top]^\top \in\mathbb{R}^6$, $e_x \coloneqq [e_p^\top, \frac{1}{2(2-e_{\scr O})^2}e_R^\top R_{\scr O}^\top]^\top$,  $K_p\coloneqq \textup{diag}\{ K_{p_1}, k_{p_2} I_3 \}$, where $K_{p_1} \in\mathbb{R}^{3\times 3}, K_d\in\mathbb{R}^{6\times 6}$ are positive definite matrices, and $k_{p_2}\in\mathbb{R}_{>0}$ is a positive constant.
		Then the solution of the closed-loop coupled system satisfies the following:
		\begin{enumerate}			
			\item $e_{\scr O}(t) < 2$, $\forall t\in\mathbb{R}_{\geq 0}$
			\item $\lim_{t\to\infty}(p_{\scr O}(t) - p_\textup{d}(t) = 0_3$,  $\lim_{t\to\infty} R_\textup{d}(t)^\top R_{\scr O}(t) =  I_3$
			\item It holds that $h_\textup{int}(t) = 0 \Leftrightarrow G^\ast = MG^\top(GMG^\top)^{-1}$.
		\end{enumerate}   
	\end{corollary}
}
    {\begin{IEEEproof} 
				1)	By substituting \eqref{eq:u rot mat} in \eqref{eq:coupled dynamics} and using $G G^\ast = I$ and $Gu_\mathcal{R} = 0_6$, we obtain, in view of \eqref{eq:coupled dynamics} and the positive definiteness of $\widetilde{M}$ that $\widetilde{M}(\bar{x})( \dot{e}_v + K_d e_v + K_p e_x )$ implying
				\begin{align}
					&\dot{e}_v = -K_d e_v - K_p e_x. \label{eq:dot_e_v closed loop}
				\end{align}			
				{Consider now the function} $V \coloneqq \frac{1}{2}e_p^\top K_{p_1} e_p + \frac{k_{p_2}}{2-e_{\scr O}} + \frac{1}{2}e_v^\top e_v,$
				for which it holds $V(0) < \infty$, since $e_{\scr O}(0) < 2$. By differentiating $V$, and using \eqref{eq:e_O_dot rot mat} and \eqref{eq:dot_e_v closed loop}, one obtains $\dot{V} = -e_v^\top K_d e_v \leq 0$.
				Hence, it holds that $V(t) \leq V(0) < \infty$, which implies that $\frac{k_{p_2}}{2-e_{\scr O}(t)}$ is bounded and consequently $e_{\scr O}(t) < 2$. \\
				2) Since $V(t) \leq V(0) < \infty$, the errors $e_p$, $e_v$ are bounded, 	which, given the boundedness of the desired trajectories $p_\textup{d}$, $R_\textup{d}$ and their derivatives, implies the boundedness of the control law $u$. Hence, it can be proved that $\ddot{V}$ is bounded which implies the uniform continuity of $\dot{V}$. Therefore, according to
				Barbalat's lemma (\cite{khalil1996noninear}, Lemma 8.2), we deduce that $\lim_{t\to\infty} \dot{V}(t) = 0 \Rightarrow \lim_{t\to\infty} e_v(t) = 0_6$. Since $e_x(t)$ is also bounded, it can be proved by using the same arguments that $\lim_{t\to\infty} \dot{e}_v(t) = 0_6$ and hence \eqref{eq:dot_e_v closed loop} implies that $\lim_{t\to\infty}e_x(t) = 0_6$. \\ 	
				3) {Let the desired object force be 
					\begin{equation} \label{eq:h O des}
						h_{\scr O,\text{d}} = C_{\scr O}v_{\scr O} + g_{\scr O}  + M_{\scr O}\alpha_\textup{d},
					\end{equation}
					where $\alpha_\textup{d} \coloneqq \dot{v}_\textup{d} - K_d e_v - K_p e_x$. 
					In view of Th. \ref{th:optimal internal force distribution}, it suffices to prove $h = h_{\text{d}} = G^\ast h_{\scr O,\text{d}}$. By substituting \eqref{eq:u rot mat} in \eqref{eq:h 1} and canceling terms, we obtain
					\begin{align*}
						h =& (M^{-1} + G^\top M_{\scr O}^{-1}G)^{-1}[M^{-1}G^\ast h_{\scr O,\textup{d}} + G^\top \alpha_\textup{d} + \\
						& G^\top M_{\scr O}^{-1}(C_{\scr O}v_{\scr O} + g_{\scr O}) ].
					\end{align*}
					Next, we add and subtract the term $G^\top M_{\scr O}^{-1} GG^\ast h_{\scr O,\text{d}}$ as
					\begin{align*}
						h =& (M^{-1} + G^\top M_{\scr O}^{-1} G)^{-1}(M^{-1} + G^\top M_{\scr O}^{-1} G)G^\ast h_{\scr O,\text{d}} + \\
						& (M^{-1} + G^\top M_{\scr O}^{-1} G)^{-1}[G^\top M_{\scr O}^{-1}(M_{\scr O}\alpha_\text{d} + C_{\scr O}v_{\scr O} + \\
						&g_{\scr O} - G^\top M_{\scr O} h_{\scr O,\text{d}} ) ],
					\end{align*}
					which, in view of \eqref{eq:h O des}, becomes $h = G^\ast h_{\scr O,\text{d}}$. Completion of the proof follows by invoking Th. \ref{th:optimal internal force distribution}. }
			\end{IEEEproof} }


		In case it is required to achieve a \textit{desired} internal force $h_\textup{int,d}$, one can add in \eqref{eq:u rot mat} a term of the form described next.

		\begin{corollary} \label{corol:h int d}
			Let $h_\textup{int,d} \in \textup{null}(G)$ be a desired internal force to be achieved. Then adding the extra term $u_\textup{int,d} = (I_{6N} - MG^\top(GMG^\top)^{-1}G)h_\textup{int,d}$  
			in  \eqref{eq:u rot mat} achieves $h_\textup{int} = h_\textup{int,d}$.
		\end{corollary}
		{ 
		\begin{IEEEproof}[Proof of Corollary \ref{corol:h int d}]
				Since $h_{\text{int,d}} \in \textup{null}(G) = \textup{range}(\mathcal{R}_\mathcal{G}^\top)$, it holds that $M^{-\frac{1}{2}}h_{\text{int,d}} \in \textup{range}(M^{-\frac{1}{2}}\mathcal{R}_\mathcal{G}^\top) = \textup{range}(\mathcal{R}_\mathcal{G}M^{-\frac{1}{2}})^\dagger$. Therefore, it holds that 
				\begin{align} \label{eq:des int 1}
					(\mathcal{R}_\mathcal{G}M^{-\frac{1}{2}})^\dagger \mathcal{R}_\mathcal{G} M^{-1}h_{\text{int,d}} =& \notag \\
					(\mathcal{R}_\mathcal{G}M^{-\frac{1}{2}})^\dagger \mathcal{R}_\mathcal{G} M^{-\frac{1}{2}} (M^{-\frac{1}{2}}h_{\text{int,d}}) =& M^{-\frac{1}{2}}h_{\text{int,d}}.
				\end{align}
				Hence, \eqref{eq:internal forces 1_2} yields the resulting internal forces
				\begin{align*}
					h_{\text{int}} =& M^{\frac{1}{2}} (\mathcal{R}_\mathcal{G} M^{-\frac{1}{2}})^\dagger \mathcal{R}_\mathcal{G} M^{-1}(I - MG^\top (GMG^\top)^{-1}G)h_{\text{int,d}} \\
					=& M^{\frac{1}{2}} (\mathcal{R}_\mathcal{G}M^{-\frac{1}{2}})^\dagger \mathcal{R}_\mathcal{G}M^{-1}h_{\text{int,d}} 
					= M^{\frac{1}{2}} M^{-\frac{1}{2}} h_{\text{int,d}} = h_{\text{int,d}},
				\end{align*}
				where we have used \eqref{eq:des int 1} and the fact that $\mathcal{R}_\mathcal{G}G^\top = 0$. 
			\end{IEEEproof}		
			}

	\begin{figure}
		\centering
		\includegraphics[width = 0.25\textwidth]{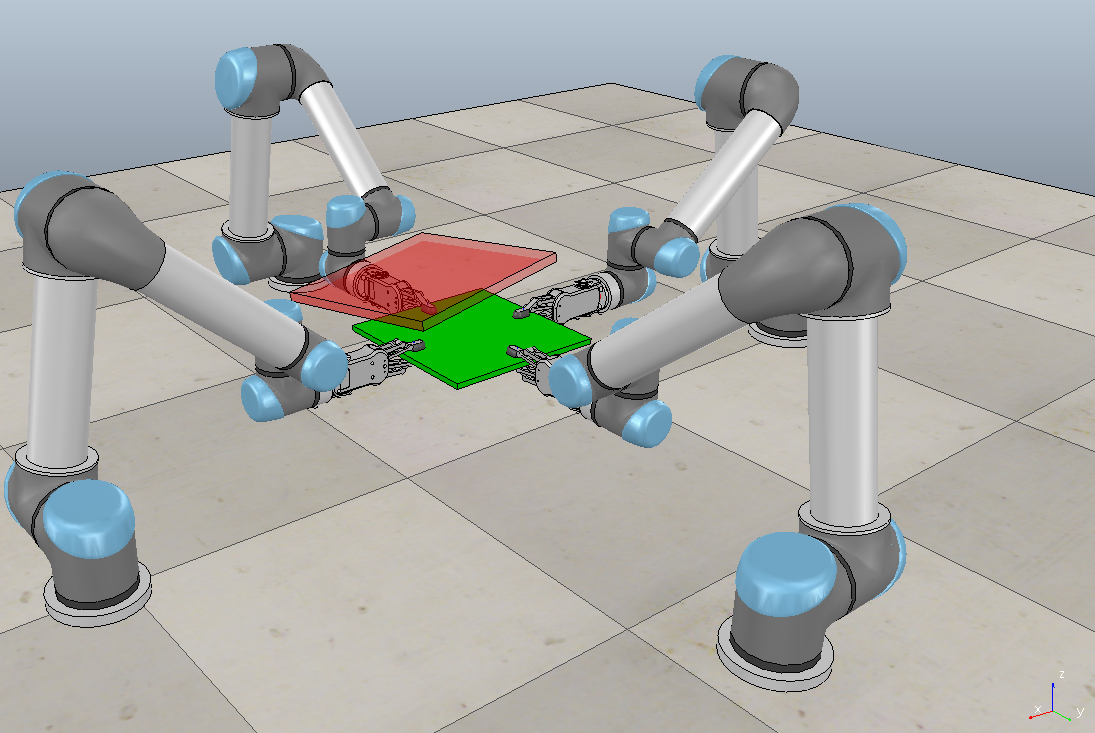}
		\caption{Four UR$5$ robotic arms rigidly grasping an object. The red counterpart represents a desired object pose at $t=0$.\label{fig:vrep_initial}}
	\end{figure}

	\section{{Discussion}} \label{sec:discussion}
		  {We briefly comment now on some of the features of the aforementioned analysis.}
		  
		 {Firstly, we note that the aforementioned results, spanning from Theorem \ref{th:null G range R_T} to Corollary \ref{corol:h int d}, still hold if the rigidity matrix $\mathcal{R}_{\mathcal{G}}$ is replaced by any constraint matrix $\mathcal{A}$ satisfying $\mathcal{A}v = 0$ and \eqref{eq:A_int, b_int}, not necessarily related to rigidity theory (see, e.g., eq. (10) of \cite{verginis_submitted}). However, the connection between cooperative manipulation and rigidity theory could, through the insight provided by Theorem \ref{th:null G range R_T}, pave the way for drawing novel links that could help solve problems in the two fields. For instance, one could use rigidity-theory results towards the localization of robotic agents \cite{zhao2016localizability}, or object-agent contact maintenance via means of rigidity maintenance \cite{zelazo2015decentralized} for non-rigid grasping contacts. Reversely, the association of cooperative manipulation and rigidity theory could help use cooperative-manipulation tools to tackle issues in formations of rigid graphs.  For instance, the results on cooperative manipulation free from internal forces could be used to prevent local minima in decentralized formation control of rigid graphs through Theorem \ref{th:null G range R_T}.
		 }
			
		Secondly, note that $G^\star = MG^\top(GMG^\top)^{-1}$ induces an \textit{implicit} and natural load-sharing scheme via the incorporation of $M$. More specifically, note that the force distribution to the robotic agents via $G^\ast h_{\scr O,\textup{d}}$ yields for each agent $M_i J_{\scr O_i} (\sum_{i\in\mathcal{N}} J_{\scr O_i}^\top M_i J_{\scr O_i})^{-1}$, $\forall i\in\mathcal{N}$. Hence, larger values of $M_i$ will produce larger inputs for agent $i$, implying that agents with larger inertia characteristics will take on a larger share of the object load.
		Note that this is also a \textit{desired} load-sharing scheme, since larger dynamic values usually imply more powerful robotic agents. Previous works (e.g., \cite{tsiamis2015cooperative}) used load-sharing coefficients, without relating the resulting force distribution with the arising internal forces.
		
		Thirdly, note that the employed controller requires knowledge of the agent and object dynamics. In case of dynamic parameter uncertainty}, standard adaptive control schemes that attempt to estimate potential uncertainties in the model (e.g., \cite{verginis2019Robust,marino2017distributed}) would intrinsically create internal forces, since the dynamics of the system would not be accurately compensated. The same holds for schemes that employ force/torque sensors that provide the respective measurements at the grasp points (e.g., \cite{tsiamis2015cooperative,heck2013internal}) in periodic time instants. Since the interaction forces depend explicitly on the control input, such measurements will unavoidably correspond to the interaction forces of the previous time instants due to causality reasons, creating thus small disturbances in the dynamic model. 
		
	    {Finally, note that the aforementioned results do not hold in degenerate cases where the rigidity matrix loses rank (see Sec. \ref{sec:Rigidity}). In such cases, $\textup{null}(\mathcal{R}_\mathcal{G})$ contains more motions than the trivial ones (coordinated translations and rotations), and hence the constraints \eqref{eq:A_int, b_int} are not consistent with the motion of the cooperative manipulation system, leading to {an inaccurate expression in} \eqref{eq:internal forces 1}. In these cases, one can employ the constraints' matrix used in \cite{verginis_submitted} (see eq. (10)), whose nullspace always coincides with $\textup{range}(G^\top)$.}

	\section{Simulation Results} \label{sec:Sim/Exp results}
	{This section provides simulation results using $4$ identical UR5 robotic manipulators in the realistic dynamic environment V-REP \cite{Vrep}. The  agents are rigidly grasping an object of $40$ kg as shown in Fig. \ref{fig:vrep_initial}. In order to verify the findings of the previous sections, we apply the controller \eqref{eq:u rot mat} to achieve tracking of a desired trajectory by the object's center of mass. We simulate the closed loop system for two cases of $G^\ast$, namely the proposed one $G^\ast_1 = MG^\top(GMG^\top)^{-1}$ as well as the more standard choice $G^\ast_2 = G^\top(GG^\top)^{-1}$, showing the validity of Coroll. \ref{corol:u design rot mat} and \ref{corol:h int d}.}
	
	{The initial pose of the object is set as $p_{\scr O}(0) = [-0.225,-0.612,0.161]^\top$, $\eta_{\scr O}(0) = [0,0,0]^\top$ and the desired trajectory as $p_\text{d}(t) = p_{\scr O}(0) + [0.2\sin(w_p{d}t + \varphi_\text{d}), 0.2\cos(w_pt+\varphi_\text{d}), 0.09+0.1\sin(w_pt + \varphi_\text{d})]^\top$, $\eta_\text{d}(t) = [0.15\sin(w_\phi t + \varphi_\text{d}),0.15\sin(w_\theta t + \varphi_\text{d}), 0.15\sin(w_\psi t + \varphi_\text{d})]^\top$ (in meters and rad, respectively), where $\varphi_\text{d} = \frac{\pi}{6}$, $w_p = w_\phi = w_\psi = 1$, $w_\theta = 0.5$, and $\eta_\text{d}(t)$ is transformed to the respective $R_\text{d}(t)$. The gains are set as $K_{p_1} = 15$, $k_{p_2} = 75$, $K_d = 40 I_6$.}
	
	{
		The results are given in {Figs. \ref{fig:errors}-\ref{fig:inputs_norm}} for $15$ seconds. Fig. \ref{fig:errors} depicts the pose and velocity errors $e_p(t)$, $e_{\scr O}(t)$, $e_v(t)$, which converge to zero for both choices of $G^\ast$, as expected. The norms of the control inputs $\tau_i(t)$ of the agents are shown in Fig. \ref{fig:inputs_norm}. Moreover,		
		the norm of the internal forces, $\|h_{\text{int}}(t)\|$, computed via \eqref{eq:internal forces 1_1}, is shown in Fig. \ref{fig:h_int} (left). It is clear that $G^\ast_2$ yields significantly larger internal forces, whereas  $G^\ast_1$ keeps them very close to zero, as proven in the theoretical analysis. The larger internal forces in the case of $G^\ast_2$ are associated with the larger control inputs $\tau_i$. This can be also concluded from Fig. \ref{fig:inputs_norm}; It is clear that inputs of larger magnitude occur in the case of $G^\ast_2$, which create internal forces. 
		}
	
	\begin{figure}
		\centering
		\includegraphics[width = 0.5\textwidth]{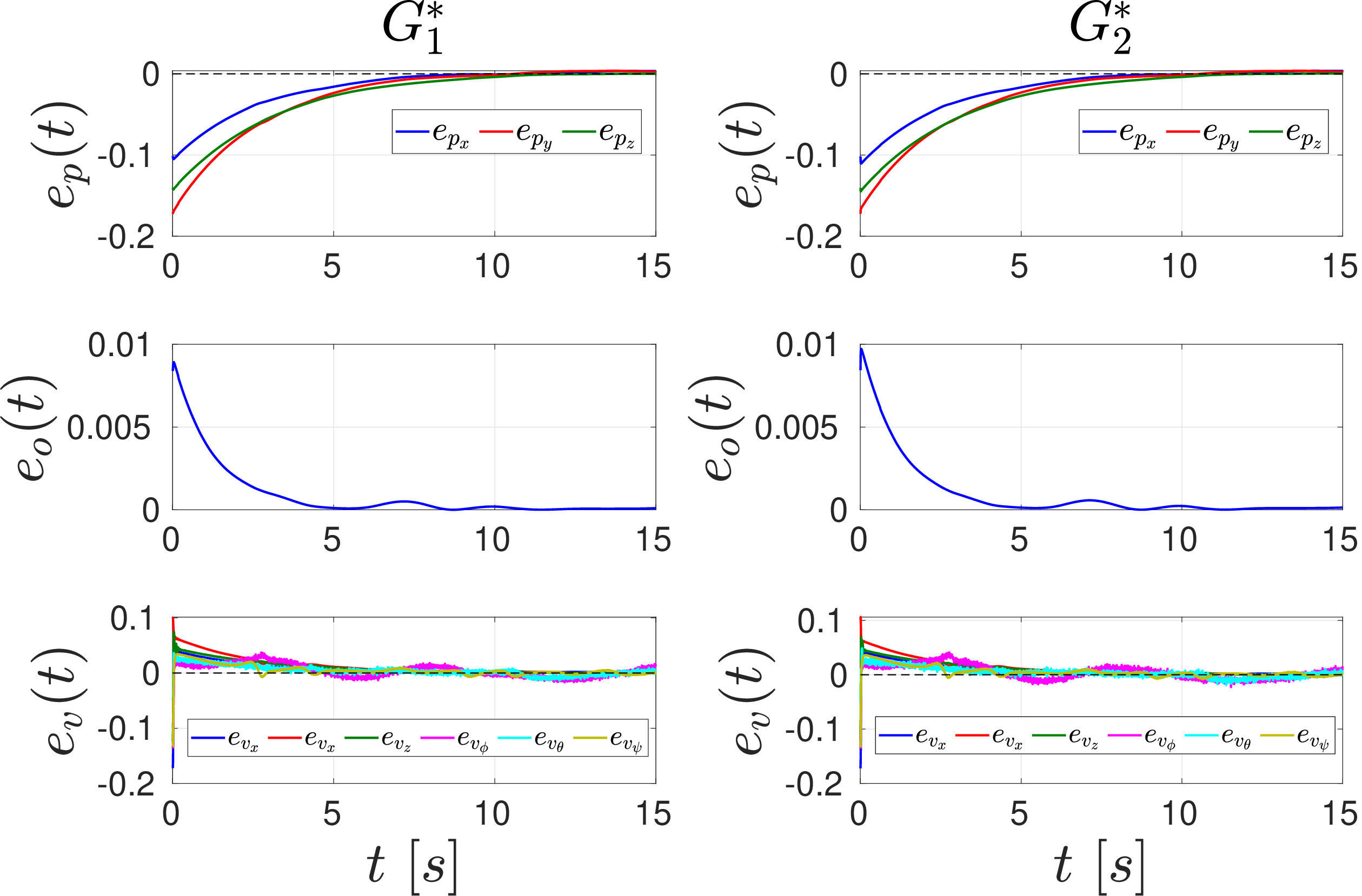}
		\caption{The error metrics $e_p(t)$, $e_{\scr O}(t)$, $e_v(t)$, respectively, top to bottom, for the two choices $G^\ast_1$ and $G^\ast_2$ and $t\in[0,15]$ seconds.\label{fig:errors}}	
	\end{figure}
		
		
		\begin{figure}
			\centering
			\includegraphics[width = 0.4\textwidth]{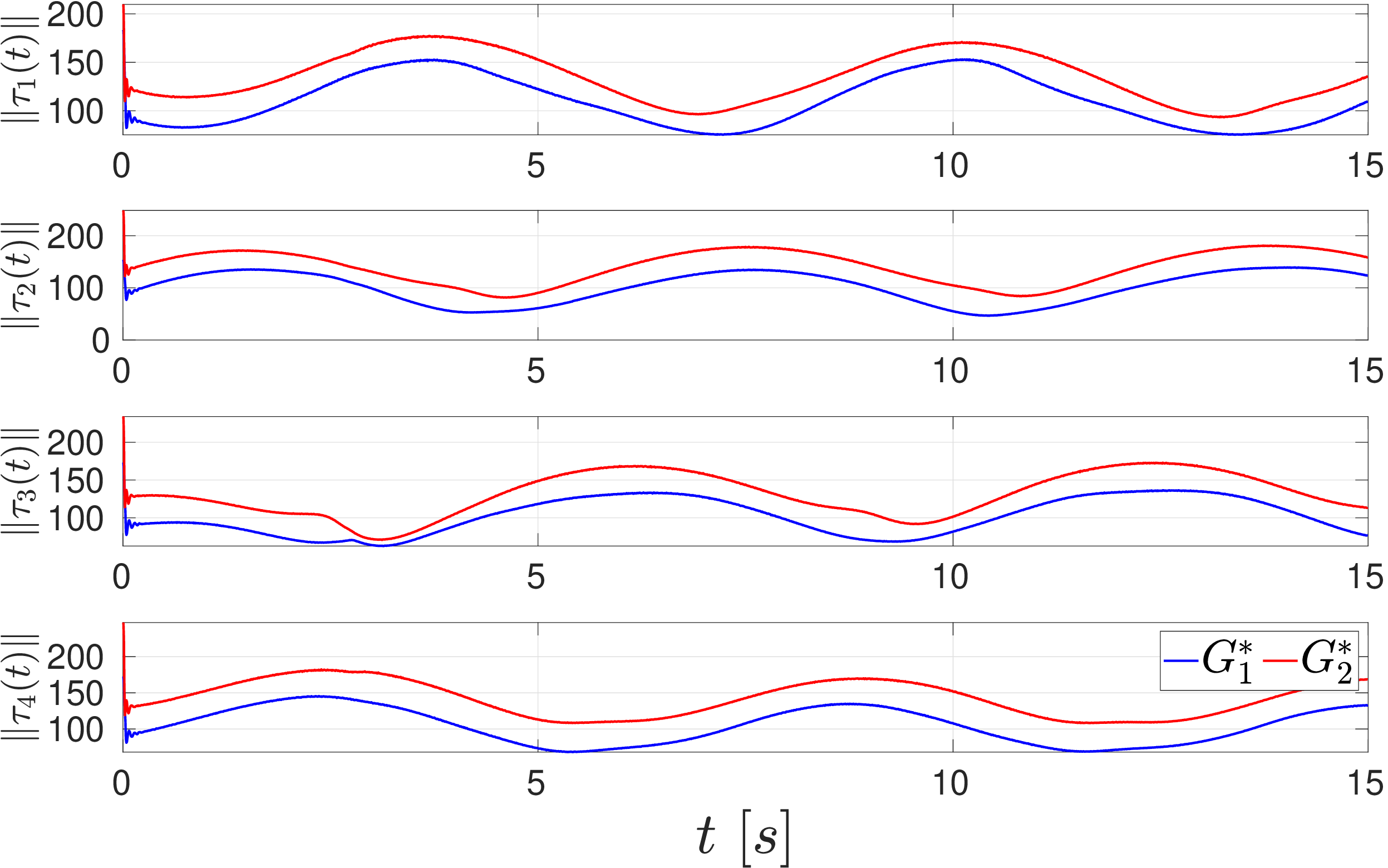}
			\caption{The norms of the resulting control inputs, $\|\tau_i(t)\|$ for $G^\ast_1$ (with blue) and $G^\ast_2$ (with red), $\forall i\in\{1,\dots,4\}$, and $t\in[0,15]$ seconds.\label{fig:inputs_norm}}
		\end{figure}
		
	\begin{figure}
		\centering
		\includegraphics[width = 0.24\textwidth]{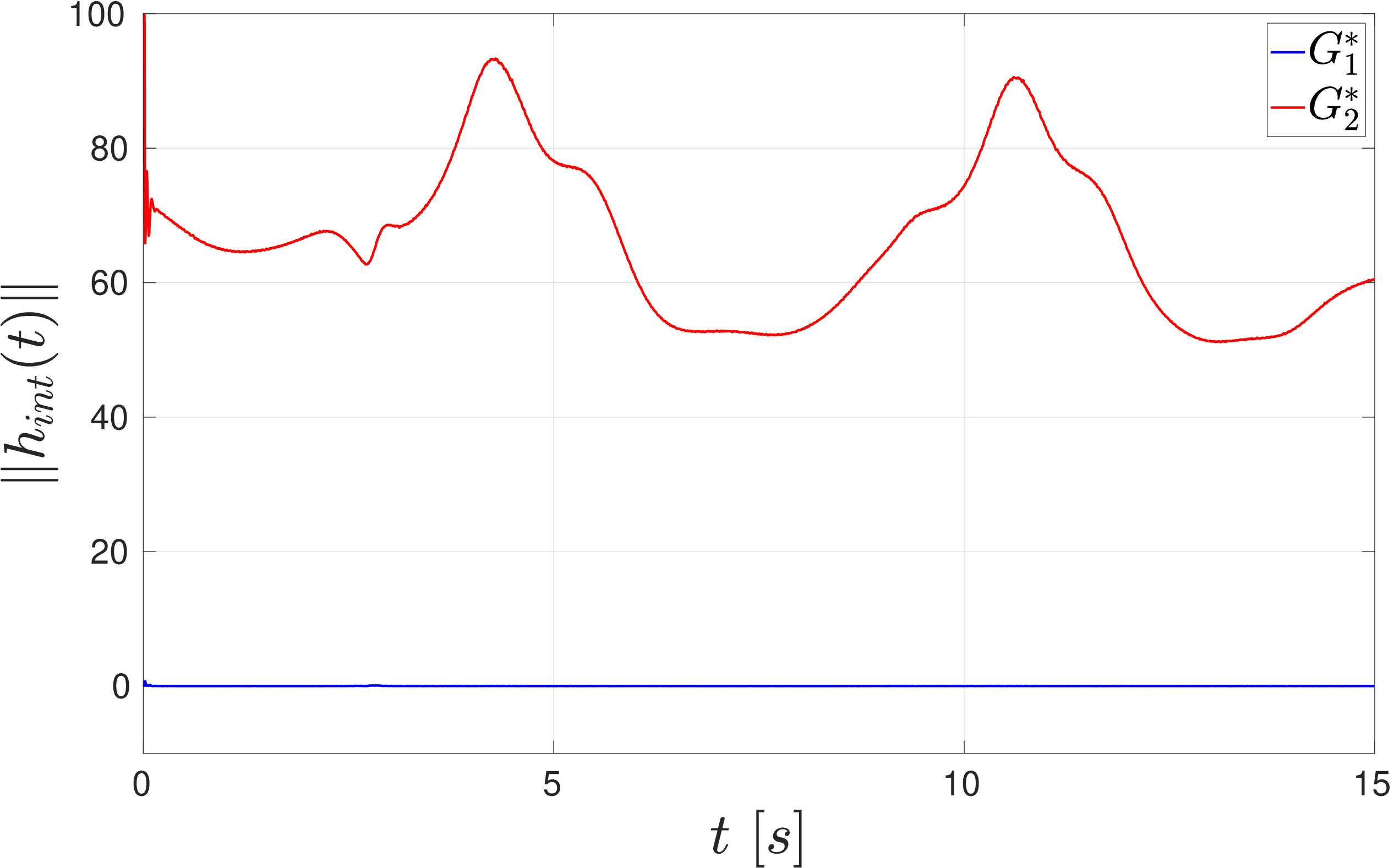}
		\includegraphics[width = 0.24\textwidth]{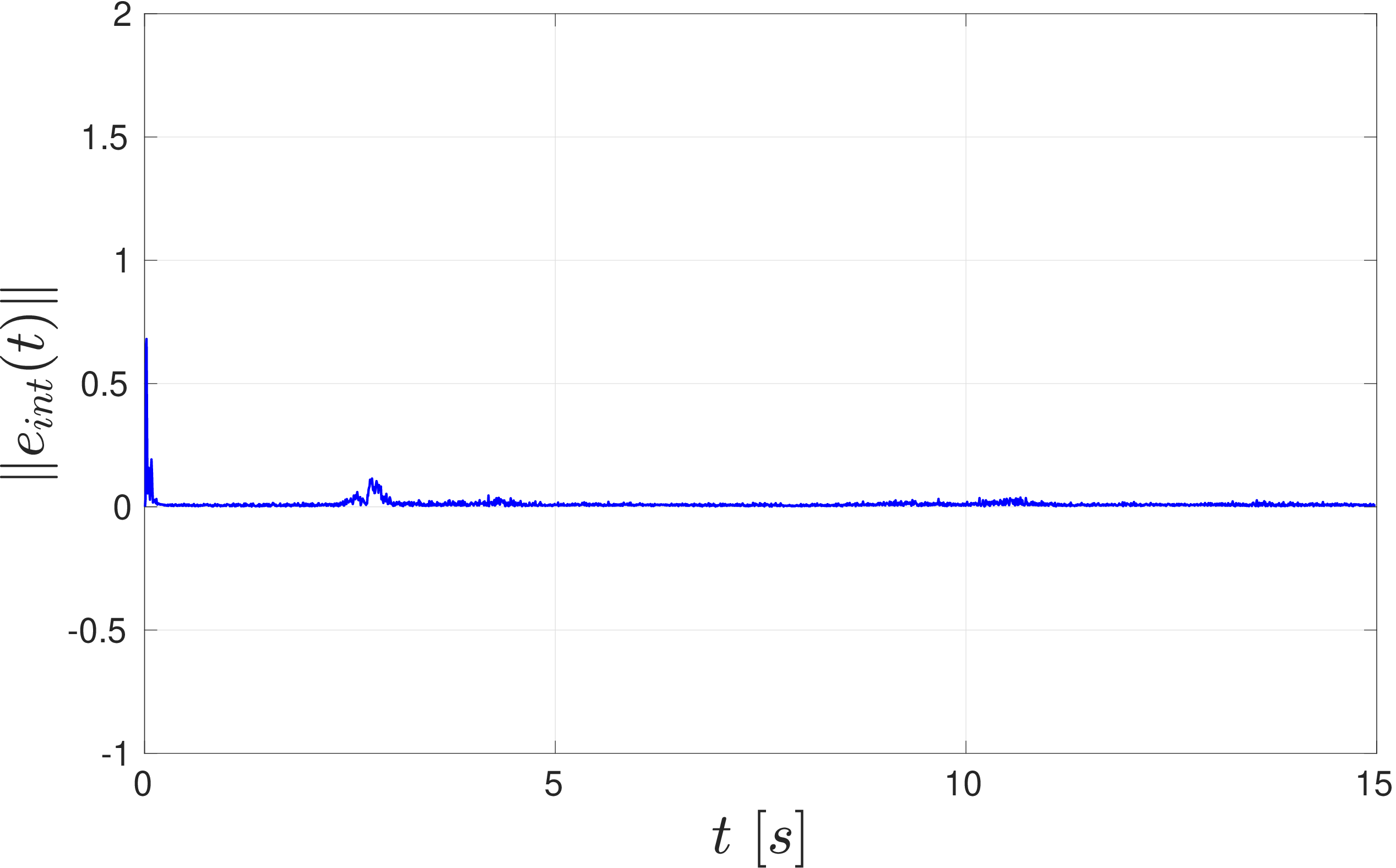}
		\caption{Left: The signal $\|h_\text{int}(t)\|$ (as computed via \eqref{eq:internal forces 1_1}) for the two cases of $G^\ast$ and $t\in[0,15]$ seconds. Right: The signal $\|e_{\text{int}}(t)\| $, when using $G^\ast_1$ and for $t\in[0,15]$ seconds. \label{fig:h_int}}
	\end{figure}

	{Finally, we set a random force vector $h_{\text{int,d}}$ in the nullspace of $G$ and we simulate the control law \eqref{eq:u rot mat} with the extra component $u_{\text{int,d}} = h_{\text{int,d}}$ (see Coroll. \ref{corol:h int d}{)}. Fig. \ref{fig:h_int} (right) illustrates the error norm $\|e_{\text{int}}(t)\|\coloneqq \|h_{\text{int,d}}(t) - h_{\text{int}}(t)\|$, which evolves close to zero. The minor observed deviations can be attributed to model uncertainties and hence the imperfect cancellation of the respective dynamics via \eqref{eq:u rot mat}. }
	

	\section{Conclusion and Future Work} \label{sec:Conclusion+FW}
	We introduce the notion of distance and bearing rigidity in $\mathsf{SE}(3)$ and we use the associated rigidity matrix
	to express the internal forces that emerge in a cooperative manipulation scheme. Based on these results, we connect the rigidity and grasp matrices via a nullspace-range relation and we provide novel results on internal-forced based cooperative manipulation control and on the relation between the interaction and internal forces. Future efforts will be directed towards using rigidity theory for object pose estimation and robust control design that minimizes the arising internal forces.

	
	%




	\ifCLASSOPTIONcaptionsoff
	\newpage
	\fi
	
	\bibliographystyle{IEEEtran}
	\bibliography{bibl}

\begin{thebibliography}{10}
\providecommand{\url}[1]{#1}
\csname url@samestyle\endcsname
\providecommand{\newblock}{\relax}
\providecommand{\bibinfo}[2]{#2}
\providecommand{\BIBentrySTDinterwordspacing}{\spaceskip=0pt\relax}
\providecommand{\BIBentryALTinterwordstretchfactor}{4}
\providecommand{\BIBentryALTinterwordspacing}{\spaceskip=\fontdimen2\font plus
\BIBentryALTinterwordstretchfactor\fontdimen3\font minus
  \fontdimen4\font\relax}
\providecommand{\BIBforeignlanguage}[2]{{%
\expandafter\ifx\csname l@#1\endcsname\relax
\typeout{** WARNING: IEEEtran.bst: No hyphenation pattern has been}%
\typeout{** loaded for the language `#1'. Using the pattern for}%
\typeout{** the default language instead.}%
\else
\language=\csname l@#1\endcsname
\fi
#2}}
\providecommand{\BIBdecl}{\relax}
\BIBdecl

\bibitem{oh2015survey}
K.-K. Oh, M.-C. Park, and H.-S. Ahn, ``A survey of multi-agent formation
  control,'' \emph{Automatica}, vol.~53, pp. 424--440, 2015.

\bibitem{Automatica_formation_18}
C.~K. Verginis, A.~Nikou, and D.~V. Dimarogonas, ``Robust formation control in
  se(3) for tree-graph structures with prescribed transient and steady state
  performance,'' \emph{Automatica}, vol. 103, pp. 538--548, 2018.

\bibitem{de2016distributed}
H.~G. De~Marina, B.~Jayawardhana, and M.~Cao, ``Distributed rotational and
  translational maneuvering of rigid formations and their applications,''
  \emph{Transactions on Robotics}, vol.~32, no.~3, pp. 684--697, 2016.

\bibitem{chen2017global}
X.~Chen, M.-A. Belabbas, and T.~Ba{\c{s}}ar, ``Global stabilization of
  triangulated formations,'' \emph{SIAM Journal on Control and Optimization},
  vol.~55, no.~1, pp. 172--199, 2017.

\bibitem{zelazo2015decentralized}
D.~Zelazo, A.~Franchi, H.~H. B{\"u}lthoff, and P.~Robuffo~Giordano,
  ``Decentralized rigidity maintenance control with range measurements for
  multi-robot systems,'' \emph{The International Journal of Robotics Research},
  vol.~34, no.~1, pp. 105--128, 2015.

\bibitem{tian2013global}
Y.-P. Tian and Q.~Wang, ``Global stabilization of rigid formations in the
  plane,'' \emph{Automatica}, vol.~49, no.~5, pp. 1436--1441, 2013.

\bibitem{zhao2019bearingRig}
S.~{Zhao} and D.~{Zelazo}, ``Bearing rigidity theory and its applications for
  control and estimation of network systems: Life beyond distance rigidity,''
  \emph{IEEE Control Systems Magazine}, vol.~39, pp. 66--83, 2019.

\bibitem{Tron15RigidComp}
R.~{Tron}, L.~{Carlone}, F.~{Dellaert}, and K.~{Daniilidis}, ``Rigid components
  identification and rigidity control in bearing-only localization using the
  graph cycle basis,'' \emph{American Control Conference}, pp. 3911--3918,
  2015.

\bibitem{lin2018projected}
H.-C. Lin, J.~Smith, K.~K. Babarahmati, N.~Dehio, and M.~Mistry, ``A projected
  inverse dynamics approach for multi-arm cartesian impedance control,''
  \emph{Intern. Conf. Robot. Autom.}, pp. 1--5, 2018.

\bibitem{aghili2005unified}
F.~Aghili, ``A unified approach for inverse and direct dynamics of constrained
  multibody systems based on linear projection operator: applications to
  control and simulation,'' \emph{IEEE Transactions on Robotics}, vol.~21,
  no.~5, pp. 834--849, 2005.

\bibitem{erhart2015internal}
S.~Erhart and S.~Hirche, ``Internal force analysis and load distribution for
  cooperative multi-robot manipulation,'' \emph{Transactions on Robotics},
  2015.

\bibitem{Udwadia92NewPerspective}
F.~Udwadia and R.~E.~Kalaba, ``A new perspective on constrained motion,''
  \emph{Proceedings of The Royal Society A: Mathematical, Physical and
  Engineering Sciences}, vol. 439, pp. 407--410, 1992.

\bibitem{Udwadia93Constrained}
R.~E.~Kalaba and F.~Udwadia, ``Equations of motion for nonholonomic,
  constrained dynamical systems via gauss’s principle,'' \emph{ASME. J. Appl.
  Mech}, vol.~60, no.~3, pp. 662--668, 1993.

\bibitem{donner2018physically}
P.~Donner, S.~Endo, and M.~Buss, ``Physically plausible wrench decomposition
  for multieffector object manipulation,'' \emph{IEEE Transactions on
  Robotics}, vol.~34, no.~4, pp. 1053--1067, 2018.

\bibitem{khatib1996decentralized}
O.~Khatib, K.~Yokoi, K.~Chang, D.~Ruspini, R.~Holmberg, and A.~Casal,
  ``Decentralized cooperation between multiple manipulators,'' \emph{IEEE
  International Workshop on Robot and Human Communication}, 1996.

\bibitem{ficuciello2014cartesian}
F.~Ficuciello, A.~Romano, L.~Villani, and B.~Siciliano, ``Cartesian impedance
  control of redundant manipulators for human-robot co-manipulation,''
  \emph{Proceedings of the IEEE/RSJ International Conference on Intelligent
  Robots and Systems (IROS)}, pp. 2120--2125, 2014.

\bibitem{yoshikawa1993coordinated}
T.~Yoshikawa and X.-Z. Zheng, ``Coordinated dynamic hybrid position/force
  control for multiple robot manipulators handling one constrained object,''
  \emph{The International Journal of Robotics Research}, 1993.

\bibitem{erhart2013impedance}
S.~Erhart, D.~Sieber, and S.~Hirche, ``An impedance-based control architecture
  for multi-robot cooperative dual-arm mobile manipulation,'' \emph{Proceedings
  of the IEEE/RSJ International Conference on Intelligent Robots and Systems
  (IROS)}, pp. 315--322, 2013.

\bibitem{tsiamis2015cooperative}
A.~Tsiamis, C.~K. Verginis, C.~P. Bechlioulis, and K.~J. Kyriakopoulos,
  ``Cooperative manipulation exploiting only implicit communication,''
  \emph{Intern. Conf. Intell. Robots and Systems}, pp. 864--869, 2015.

\bibitem{heck2013internal}
D.~Heck, D.~Kosti{\'c}, A.~Denasi, and H.~Nijmeijer, ``Internal and external
  force-based impedance control for cooperative manipulation,'' \emph{IEEE
  European Control Conference (ECC)}, pp. 2299--2304, 2013.

\bibitem{verginis2019Robust}
C.~K. {Verginis}, M.~{Mastellaro}, and D.~V. {Dimarogonas}, ``Robust
  cooperative manipulation without force/torque measurements: Control design
  and experiments,'' \emph{Transactions on Control Systems Technology}, 2019.

\bibitem{erhart2013adaptive}
S.~Erhart and S.~Hirche, ``Adaptive force/velocity control for multi-robot
  cooperative manipulation under uncertain kinematic parameters,''
  \emph{Intern. Conf. Intell. Robots and Systems}, pp. 307--314, 2013.

\bibitem{ponce2016cooperative}
A.-N. Ponce-Hinestroza, J.-A. Castro-Castro, H.-I. Guerrero-Reyes,
  V.~Parra-Vega, and E.~Olgu{\`y}n-D{\`y}az, ``Cooperative redundant
  omnidirectional mobile manipulators: Model-free decentralized integral
  sliding modes and passive velocity fields,'' \emph{International Conference
  on Robotics and Automation (ICRA)}, pp. 2375--2380, 2016.

\bibitem{marino2017distributed}
A.~Marino, ``Distributed adaptive control of networked cooperative mobile
  manipulators,'' \emph{Trans. on Control Systems Technology}, 2017.

\bibitem{walker1991analysis}
I.~D. Walker, R.~A. Freeman, and S.~I. Marcus, ``Analysis of motion and
  internal loading of objects grasped by multiple cooperating manipulators,''
  \emph{The International journal of robotics research}, 1991.

\bibitem{williams1993virtual}
D.~Williams and O.~Khatib, ``The virtual linkage: a model for internal forces
  in multi-grasp manipulation,'' \emph{International Conference on Robotics and
  Automation}, vol.~1, pp. 1025--1030, 1993.

\bibitem{chung2005analysis}
J.~H. Chung, B.-Y. Y.~W. K., and Kim, ``Analysis of internal loading at
  multiple robotic systems,'' \emph{Journal of mechanical science and
  technology}, vol.~19, no.~8, pp. 1554--1567, 2005.

\bibitem{erhart2016model}
S.~Erhart and S.~Hirche, ``Model and analysis of the interaction dynamics in
  cooperative manipulation tasks,'' \emph{Transactions on Robotics}, 2016.

\bibitem{briot2019physical}
S.~Briot and P.~R. Giordano, ``Physical interpretation of rigidity for bearing
  formations: Application to mobility and singularity analyses,'' \emph{Journal
  of Mechanisms and Robotics}, vol.~11, no.~3, 2019.

\bibitem{verginis_submitted}
C.~K. Verginis and D.~V. Dimarogonas, ``Energy-optimal cooperative manipulation
  via provable internal-force regulation,'' \emph{IEEE International Conference
  on Robotics and Automation (ICRA)}, 2020.

\bibitem{albert72Pseudoinverse}
A.~Albert, \emph{Regression and the Moore-Penrose Pseudoinverse}, 1972.

\bibitem{siciliano2010robotics}
B.~Siciliano, L.~Sciavicco, L.~Villani, and G.~Oriolo, \emph{Robotics:
  modelling, planning and control}.\hskip 1em plus 0.5em minus 0.4em\relax
  Springer Science \& Business Media, 2010.

\bibitem{asimow1978rigidity}
L.~Asimow and B.~Roth, ``The rigidity of graphs,'' \emph{Transactions of the
  American Mathematical Society}, vol. 245, pp. 279--289, 1978.

\bibitem{bearingSE2}
D.~{Zelazo}, A.~{Franchi}, and P.~R. {Giordano}, ``Rigidity theory in se(2) for
  unscaled relative position estimation using only bearing measurements,''
  \emph{2014 European Control Conference (ECC)}, pp. 2703--2708, June 2014.

\bibitem{bearingSE3}
G.~{Michieletto}, A.~{Cenedese}, and A.~{Franchi}, ``Bearing rigidity theory in
  se(3),'' \emph{Conference on Decision and Control}, pp. 5950--5955, 2016.

\bibitem{Michieletto2021}
G.~Michieletto, D.~Zelazo, and A.~Cenedese, ``A unified dissertation on bearing
  rigidity theory,'' \emph{IEEE Transactions on Control of Network Systems
  (early access)}, 2021.

\bibitem{boyd2004convex}
S.~Boyd and L.~Vandenberghe, \emph{Convex optimization}.\hskip 1em plus 0.5em
  minus 0.4em\relax Cambridge university press, 2004.

\bibitem{gould1985practical}
N.~I. Gould, ``On practical conditions for the existence and uniqueness of
  solutions to the general equality quadratic programming problem,''
  \emph{Mathematical Programming}, vol.~32, no.~1, pp. 90--99, 1985.

\bibitem{khalil1996noninear}
H.~K. Khalil, ``Noninear systems,'' \emph{Prentice-Hall, New Jersey}, vol.~2,
  no.~5, pp. 5--1, 1996.

\bibitem{zhao2016localizability}
S.~Zhao and D.~Zelazo, ``Localizability and distributed protocols for
  bearing-based network localization in arbitrary dimensions,''
  \emph{Automatica}, vol.~69, pp. 334--341, 2016.

\bibitem{Vrep}
E.~Rohmer, S.~P. Singh, and M.~Freese, ``V-rep: a versatile and scalable robot
  simulation framework,'' \emph{Proceedings of the IEEE/RSJ International
  Conference on Intelligent Robots and Systems (IROS)}, 2013.

\end{thebibliography}

\end{document}